\documentclass{goodfire}

\usepackage{goodfire}

\graphicspath{{../}{./}}

\usepackage{template/conference/arxiv}

\usepackage{template/conference/custom}

\usepackage{booktabs,longtable,array,amsmath,amssymb}
\usepackage[cal=boondox]{mathalfa}
\usepackage{textcomp}
\usepackage[most]{tcolorbox}
\usepackage{xcolor}
\usepackage{hyperref}
\usepackage{xspace}
\usepackage{wrapfig}
\usepackage{pgfplots}
\pgfplotsset{compat=1.18}

\colorlet{ember}{goodfire2}
\colorlet{slate}{slate700}
\colorlet{forest}{emerald600}

\definecolor{gfNavy}{HTML}{121B34}
\definecolor{gfOrange}{HTML}{F28C52}
\definecolor{gfDataBg}{HTML}{EDF2FA}
\definecolor{gfModelBg}{HTML}{FDF0E6}

\newtcbox{\model}{
  on line,
  boxrule=0pt,
  arc=2pt,
  boxsep=0pt,
  left=3pt,right=3pt,top=1pt,bottom=1pt,
  colback=gfModelBg,
  colframe=gfModelBg,
  fontupper=\ttfamily
}

\newtcbox{\dataset}{
  on line,
  boxrule=0pt,
  arc=3pt,
  boxsep=0pt,
  left=3pt,right=3pt,top=1pt,bottom=1pt,
  colback=gfDataBg,
  colframe=gfDataBg,
  fontupper=\ttfamily
}

\definecolor{trainableParam}{HTML}{C2410C} 

\newcommand{\trainable}[1]{\textcolor{trainableParam}{#1}}

\usepackage{xcolor}
\usepackage[most]{tcolorbox}
\definecolor{silicocream}{HTML}{F6F5F0}
\definecolor{silicoink}{HTML}{1D272A}

\newcommand{\silicofont}{\rmfamily}

\newtcbox{\silicochip}{on line, colback=silicocream, colframe=silicocream,
  boxrule=0pt, arc=2.5pt, left=2.5pt, right=3.5pt, top=1pt, bottom=1pt,
  boxsep=0pt, nobeforeafter, tcbox raise base}

\newcommand{\SilicoMark}{%
  \raisebox{-0.18ex}{\includegraphics[height=0.92em]{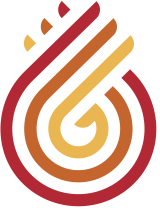}}%
}

\DeclareRobustCommand{\Silico}{%
  \texorpdfstring{\silicochip{\SilicoMark\hspace{0.12em}%
    {\silicofont\color{silicoink}Silico}}}{Silico}\xspace
}

\title{
What LLM Forecasters Know but Don't Say: \\
Probing Internal Representations for Calibration and Faithfulness
}

\authors{
Rapha\"{e}l Sarfati\textcolor{goodfire2}{~$^{\ast,g}$}, 
Pratyush Ranjan Tiwari\textcolor{goodfire2}{~$^{\ast,e}$} \\
Siddharth Boppana\textcolor{goodfire2}{~$^g$},
Christopher J. Earls\textcolor{goodfire2}{~$^g$},
Srikar Varadaraj\textcolor{goodfire2}{~$^e$},
Eric Ho\textcolor{goodfire2}{~$^g$}
}

\correspondence{core contributors: raphael@goodfire.ai, pratyush@eternis.ai}

\abstract{
Large language models fine-tuned for forecasting can be accurate yet poorly calibrated, and their chain-of-thought (CoT) reasoning may not faithfully reflect the evidence behind a forecast.
We ask whether internal representations offer a more direct window into both.
Working with \model{Eternis-Forecaster 8B} on \dataset{OpenForesight}, we train representation-pooling probes on intermediate activations and find they achieve substantially better calibration; a result that also holds for \model{GLM-4.7-Flash} and \model{GLM-4.5-Air}.
We then assess CoT faithfulness through evidence ablation and diversionary injection: 
removing an influential source in the prompt often changes the model's forecast while leaving the reasoning trace untouched. 
The same probes function as lie detectors: their activations track behavioral shifts far better than the reasoning trace does, and they also predict the direction of change in 84\% of cases, including when the CoT conceals the perturbation's influence.
Finally, forced answering reveals that forecasts are largely fixed before reasoning begins: a single pre-reasoning pass recovers the committed answer and confidence, and routing questions by the spread of this pre-set answer distribution saves 30--47\% of generated tokens, \emph{with no loss of accuracy}.
Together, these results establish probing internal representations as a practical tool for calibrating, auditing, and triaging language model forecasters and reasoning models more broadly.
}

\begin{document}

\maketitlebox

\vspace{-0.3cm}
\begin{center}
  \centering
  \includegraphics[width=0.9\linewidth]{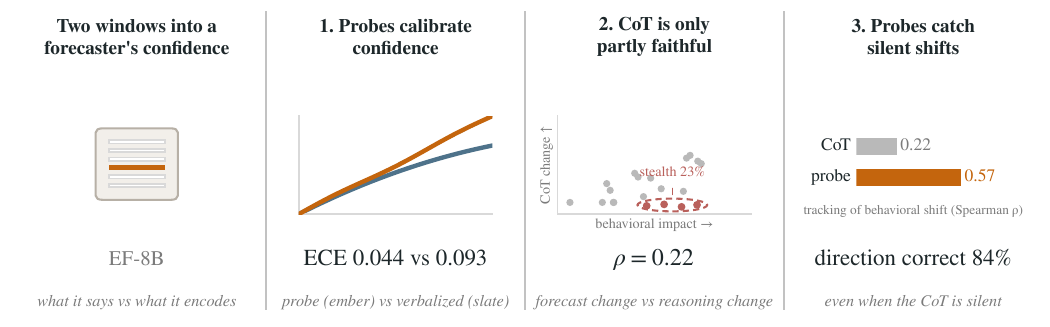}
\end{center}

\clearpage

\tableofcontents

\clearpage

\section{Introduction}
\label{sec:introduction}

\paragraph*{}
Large language models (LLMs) are, by design, next-token predictors.
They are also robust analysts and forecasters. 
Somehow the task of predicting a new token $x$ given a context $C$ has transferred the emergent ability to: reason about large amounts of textual data, recoup information, produce inferences, and finally even reason about out-of-distribution topics; thus, paving the way to new scientific discovery and robust forecasting of future events. 
Recent work has shown that LLMs, post-trained using reinforcement learning (RL), can rival or surpass human forecasters on real-world prediction markets and open-ended questions~\citep{futuresim2026,chandak2026scalingopenendedreasoningpredict}; thus, establishing them as practical tools for decision support across domains from geopolitics to public health.

\paragraph*{
Accuracy alone does not make a forecaster trustworthy. 
}
A well-calibrated forecaster must know what it knows: when it assigns 70\% confidence to an outcome, that outcome should materialize roughly 70\% of the time. 
In practice, LLM forecasters are often miscalibrated: their verbalized probability estimates deviate substantially from empirical accuracy. 
Compounding this, the chain-of-thought (CoT) reasoning that accompanies a forecast may not faithfully reflect the computation that actually produced it. 
If a model's stated reasoning omits or misrepresents the evidence driving its prediction, then the user cannot meaningfully audit the forecast, and the associated calibration derived from verbalized confidence inherits the distortions of an unreliable narrator.

\paragraph*{
In this work, we ask whether the model's internal representations offer a more direct window into both confidence and reasoning fidelity.
} 
We train lightweight probes -- linear classifiers operating on pooled intermediate activations (mean-pooling, attention-pooling, and covariance pooling) -- on the CoT representations of \model{Eternis-Forecaster 8B} (\model{EF-8B})~\citep{eternis2026sotaforecasting}, post-trained from \model{Qwen3-8B} using an approach based on reinforcement learning from verifiable rewards (RLVR) for better forecasting. 
These probes read the model's internal state to estimate the likelihood that a given forecast is correct, bypassing verbalization entirely. 
We also test probe-only calibration on two frozen models of different sizes, \model{GLM-4.7-Flash} and \model{GLM-4.5-Air}, training only probe weights on ${\sim} 10,000$ improved-context \dataset{OpenForesight} rollouts.
To test whether this signal generalizes beyond forecasting, we also run a math reasoning stress test.
We reproduce the dynamic clipping policy optimization (DCPO) recipe for \model{Qwen3-8B}~\citep{ma2026decouplingreasoningconfidenceresurrecting} and evaluate frozen correctness probes on pooled out-of-distribution (OOD) \dataset{AIME}/\dataset{AMC} benchmarks, using math as a separate calibrated-reasoning domain.
We complement this with two systematic tests of chain-of-thought faithfulness, following the causal perturbation framework of~\citet{gurarieh2026faithfulnessmetricsdontmeasure}: evidence ablation, which removes supporting articles and measures whether the reasoning updates accordingly, and diversionary injection, which inserts fabricated evidence and checks whether the model discloses its influence.

\paragraph*{
We find that internal probes substantially outperform model-verbalized confidence,
} achieving a much lower expected calibration error (ECE).
Across both \model{GLM} models, raw probe probabilities sharply reduce ECE relative to verbalized confidence, supporting calibration gains without updating model weights.
Chain-of-thought reasoning proves only partially faithful: removing influential evidence changes the forecast without updating the reasoning trace in 23\% of high-impact cases (Spearman $\rho$ = 0.22 between behavioral impact and CoT change), though the model is largely transparent when actively presented with misleading evidence (stealth rate 2.9\%). 
Most strikingly, the same probes that calibrate confidence also function as ``lie detectors'': their activations track behavioral shifts at $\rho$ = 0.57 and detect internal state changes even when the chain of thought conceals the perturbation's influence. 
\textbf{Together, these results establish internal probing as a practical tool for both calibrating and auditing language model forecasters.}

\section{Related Work}
\label{sec:related_work}

\subsection{Time series extrapolation}
A first line of work asks whether the sequence-modeling priors of LLMs transfer
to numerical time series. 
\citet{gruver2023llmtime} show that frozen, general-purpose LLMs extrapolate time series zero-shot when numeric values are
tokenized as text, rivaling purpose-built models, while \citet{jin2024timellm} reprogram a frozen LLM with learned text prototypes. 
One step further, \citet{liu2024dynamical} find that frozen \model{Llama-2} predicts the trajectories of
diverse dynamical systems in-context, uncovering an in-context neural scaling law as the evidence window grows. 
A parallel line trains dedicated time-series foundation models that borrow the tokenize-and-pretrain recipe of language modeling, including \citet{ansari2024chronos} and the decoder-only \citet{das2024timesfm}. 
The premise is contested: \citet{tan2024useful} find that ablating or replacing the LLM backbone in several such systems leaves accuracy unchanged, suggesting the gains may owe more to the surrounding architecture than to language pretraining. Our setting differs: we study forecasting of discrete real-world events from natural-language evidence, not numerical extrapolation.

\subsection{Open-ended forecasting}
Closer to our setting is judgmental forecasting of future events posed in natural
language. 
\citet{zou2022autocast} introduced Autocast, a benchmark of real tournament questions with a date-stamped news corpus, and found LLMs to be far below human expert performance levels. 
Subsequent systems narrowed the gap: 
\citet{halawi2024forecasting} pair retrieval with reasoning to approach the crowd aggregate of competitive forecasters, and \citet{schoenegger2024silicon} show an ensemble of LLMs can match human-crowd accuracy; 
\citet{karger2025forecastbench} track this progress with a contamination-free, continuously updated benchmark. 

\subsection{Calibration-aware RL for reasoning}
Reinforcement learning from verifiable rewards (RLVR) trains reasoning models with automatically checked outcomes, most often a binary correctness reward.
This improves mathematical and symbolic reasoning, but binary rewards do not distinguish confident errors from cautious guesses, so RLVR can worsen calibration even as accuracy improves.
\citet{damani2026binaryrewardstraininglms} propose Reinforcement Learning with Calibration Rewards (RLCR), which augments the correctness reward with a Brier-score calibration reward and trains the model to emit both an answer and a verbalized confidence.
The Brier term is a proper scoring rule for binary correctness, so it directly incentivizes confidence estimates that match empirical success probabilities.
\citet{ma2026decouplingreasoningconfidenceresurrecting} propose DCPO, motivated by an accuracy-calibration gradient conflict in coupled RL objectives.
DCPO separates the optimization of reasoning tokens and confidence tokens, using verbalized confidence and decoupled advantages to improve calibration while preserving the gains from RLVR.
Our math experiment asks a complementary question: after using a strong training-time calibration recipe, can a frozen activation readout still recover OOD correctness information that token statistics or verbalized confidence miss?

\section{Methods}
\label{sec:methods}
In this section, we carefully outline the foundational concepts, and associate details, that underpin the work reported on, herein.
\subsection{Models}
We focus our analysis on \model{Eternis-Forecaster 8B} (\model{EF-8B})~\citep{eternis2026sotaforecasting}, post-trained from \model{Qwen/Qwen3-8B} using an RLVR-style approach for better forecasting.
For a subset of experiments we additionally use \model{Eternis-Forecaster 32B} (\model{EF-32B}), a larger model from the same family.
As comparison baselines we also consider:
\begin{itemize}
    \item \model{nikhilchandak/OpenForecaster-8B}~\citep{chandak2026scalingopenendedreasoningpredict};
    \item the original \model{Qwen/Qwen3-8B}~\citep{yang2025qwen3technicalreport} in reasoning mode.
\end{itemize}
For the probe-only calibration generalization study, we use two frozen GLM models: \model{GLM-4.7-Flash} and \model{GLM-4.5-Air}~\citep{glmteam2025glm45}.
They differ in size, and only the probe weights are trained.

\subsection{Dataset}
\paragraph*{\textbf{Forecasting dataset.}}
We rely on the \dataset{nikhilchandak/OpenForesight} dataset, introduced in~\citet{chandak2026scalingopenendedreasoningpredict}.
The \texttt{prompt} field aggregates: initial instructions, the forecasting question itself, background and resolution criteria, and ``relevant passages from retrieved news articles''; it concludes with the following instructions:
\begin{quote}
    \texttt{Your final answer should be concise (NOT MORE THAN A FEW WORDS LONG) and your response SHOULD STRICTLY END with <answer> </answer> tags and <probability> </probability> tags.}
\end{quote}
The public \texttt{HuggingFace} release of the data used herein consists of six splits (at press time): \dataset{train} (52.2k rows), \dataset{validation} (207), \dataset{test} (302), \dataset{aljazeera2026Q1} (330), \dataset{aljazeeraLate2025} (491), and \dataset{skysports2025} (1.79k).

Additionally, we developed an improved context-building scheme and used it to derive a non-public \emph{compact-context} build of the dataset, whose prompts carry roughly half as much retrieved-article context per question.
The improved data set comprises three splits: \dataset{train\textsuperscript{*}} (47,466 rows), \dataset{test\textsuperscript{*}} (296), and \dataset{val\textsuperscript{*}} (193).
The starred \dataset{test\textsuperscript{*}} and \dataset{val\textsuperscript{*}} questions are subsets of the public \dataset{test} and \dataset{validation} splits, with identical questions, ground-truth answers, and resolution criteria but different prompts.

The forecasting models were trained on the \dataset{train} split.
The themes and format of the \dataset{aljazeera} and \dataset{skysports} splits make them out-of-distribution (OOD) relative to the original \verb|train/val/test| splits; thus, providing a broader testing environment.
A data audit (Appendix~\ref{app:data-audit}) reveals a large number of shared qids across splits (Fig.~\ref{fig:data-audit_shared-qids}), but all are collisions rather than duplicated questions (as evaluated by exact match and LLM-as-a-judge).

\paragraph*{\textbf{GLM probe-only dataset.}}
For the GLM probe-only calibration study, we use another OpenForesight version whose question contexts were improved to be more relevant than the semantic-search contexts used in the OpenForecaster paper~\citep{chandak2026scalingopenendedreasoningpredict}.
Each GLM probe is trained on $11{,}835$ rollouts from the train pool ($3{,}945$ questions $\times$ 3 completions), selected on $1{,}930$ validation rollouts, and reported on $2{,}960$ held-out test rollouts.
Probe inputs contain the prompt plus chain-of-thought truncated before the final answer block; model weights stay frozen and only probe weights are optimized with cross-entropy/BCE.
We also track leakage-controlled subsets: $423$ usable self-report rollouts for \model{GLM-4.7-Flash} after removing CoT answer leakage and $1{,}747$ rollouts for \model{GLM-4.5-Air} after removing mid-reasoning probability leakage.

\paragraph*{\textbf{Math reasoning dataset.}}
For the math reasoning calibration experiment, probes are trained on a level-stratified subset of \dataset{MATH-train}: $3{,}999$ questions after deduplicating against the math test benchmarks.
The OOD evaluation pools \dataset{AIME24} ($30 \times 4$ rollouts), \dataset{AIME25} ($30 \times 4$), \dataset{AMC23} ($40 \times 4$), and \dataset{AMC24} ($45 \times 4$), for $n=580$ rollouts.
We also track clean \dataset{MATH-500} as an in-distribution diagnostic, but the headline comparison in Section~\ref{sec:math-calibration} uses only the pooled AIME/AMC OOD split.
The DCPO-recipe probe-train pool overlaps the RLVR training distribution through \dataset{DeepScaleR}, so cross-model similarity in probe performance should not be read as evidence that the underlying representations are training-invariant.

\subsection{Probes}
\label{sec:probes}
Probes are small classifiers trained on the frozen internal activations of a larger model: they read out information the model represents internally, without modifying the model itself~\citep{alain2018understandingintermediatelayersusing,belinkov2021probingclassifierspromisesshortcomings}.

\paragraph*{Pooling probes.}
Rather than reading from a single activation~$\bm{x}_i \in \mathbb{R}^{D}$, pooling probes compute a function of the full context~$(\bm{x}_1, \dots, \bm{x}_i, \dots, \bm{x}_N)$ through a pooling function $\phi$.
Note that \emph{individual activations already contain contextual information from preceding tokens}, but that information is compressed and filtered; pooling across the sequence recovers signal that no single position retains.
Given a matrix of such activations $\bm{X} \in \mathbb{R}^{N \times D}$, a probe is defined as $f_{\trainable{\theta}}(\bm{X}) = g_{\trainable{\theta}}(\phi_{\trainable{\theta}}(\bm{X}))$, where $g_{\trainable{\theta}}$ is a readout function (typically a linear layer, see below) and $\phi_{\trainable{\theta}}$ is a pooling function; $\trainable{\theta}$ denotes a set of trainable parameters.

\paragraph{Mean-pooling} implements a naive arithmetic average along the token axis; in matrix form ($\bm{1}_{ij} = 1 \quad \forall (i,j)$):
\begin{equation}
    \phi_\mathrm{mean}(\bm{X}) = \frac{1}{N} \bm{X}^\top \bm{1}_N \, \in \mathbb{R}^D.
\end{equation}

\paragraph{Attention probes} (more precisely, \textit{attention-pooling probes}) use trained attention heads to aggregate residual-stream activations~\citep{kantamneni2025sparseautoencodersusefulcase,boppana2026reasoningtheaterdisentanglingmodel}:
\begin{equation}
    \phi_\mathrm{attn}(\bm{X}) = \bm{X}^\top \operatorname{softmax} \left( \bm{X} \trainable{\bm{q}} \right) \, \in \mathbb{R}^D,
\end{equation}
where the softmax is taken over token positions and $\trainable{\bm{q}} \in \mathbb{R}^D$ is a trained query projection.

\paragraph{Covariance probes} were introduced in~\citet{dooms2026covariance,Pearce2026.04.10.717844}. 
They compress the sequence's (uncentered) covariance structure into a low-dimensional bilinear bottleneck (dim = $k$):
\begin{equation}
    \phi_\mathrm{cov}(\bm{X}) = \frac{1}{N} \left(\bm{X} \trainable{\bm{L}}\right)^\top \left(\bm{X} \trainable{\bm{R}} \right) \, \in \mathbb{R}^{k \times k},
\end{equation}
where $\trainable{\bm{L}}$ and $\trainable{\bm{R}} \in \mathbb{R}^{D \times k}$ are trained low-rank projections; the resulting matrix is flattened and passed to the readout.
Unlike mean- and attention-pooling, this captures second-order (co-fluctuation) statistics of the activations across the sequence.

\paragraph*{Readout function.} 
The readout function $g_{\trainable{\theta}}$ is generally a simple scalar product; in our case:
\begin{equation}
    f_{\trainable{\theta}}(\bm{X}) = \langle \trainable{\bm{W}}, \phi_{\trainable{\theta}}(\bm{X}) \rangle + \trainable{b},
\end{equation}
where $\langle \cdot , \cdot \rangle$ is the appropriate inner product: the vector dot product for mean and attention pooling, and the Frobenius inner product for covariance pooling.

\subsection{Metrics}
\label{sec:metrics}
We use standard statistical metrics to evaluate prediction calibration~\citep{damani2026binaryrewardstraininglms}.
A prediction~$\pi$ is defined as a triplet consisting of an answer $a$, an associated confidence $p \in [0,1]$, and an outcome $y \in \{0,1\}$ indicating correctness: $\pi = \left( a, p, y\right)$.

\paragraph*{Expected Calibration Error (ECE).}
ECE measures the agreement between stated confidence and empirical accuracy:
\textit{when the model says it is 40\% confident, is it correct 40\% of the time?}
Confidence values are binned and compared against the empirical accuracy within each bin:
\begin{equation}
    \mathrm{ECE} = \sum_b \frac{n_b}{N} \left\vert \mathrm{acc}(b) - \mathrm{conf}(b) \right\vert,
\end{equation}
where $n_b$ is the number of predictions in bin $b$, $N$ the total number of predictions, $\mathrm{acc}(b)$ the fraction of correct predictions in the bin, and $\mathrm{conf}(b)$ the mean confidence in the bin.

\paragraph*{Brier score.}
This measures the cost of misplaced confidence: \textit{does the model pay for its confidence when it is wrong?}
Given $N$ predictions with confidences $p_i \in [0,1]$ and outcomes $y_i$, the Brier score is the mean squared error between the two:
\begin{equation}
    \mathrm{Br} = \frac{1}{N} \sum_i \left( p_i - y_i\right)^2.
\end{equation}
It rewards confident correct predictions and penalizes confident errors quadratically.

\paragraph*{AUROC.}
Calibration and discrimination are complementary properties of a confidence signal.
ECE and the Brier score ask whether confidence values \emph{match} empirical accuracy; 
Area Under the Receiver Operating Characteristic curve (AUROC) asks whether confidence \emph{separates} correct predictions from incorrect ones.
Imagine drawing one correct and one incorrect prediction at random and betting that the correct one carries higher confidence: AUROC is simply how often you would win that bet (0.5 = no better than a coin flip, 1 = the confidence always ranks the correct answer above the incorrect one).
Equivalently, it sweeps every possible confidence threshold and measures how cleanly the two groups can be told apart, which is why it needs no threshold to be chosen in advance.
Because it depends only on the ordering of confidences, not their absolute values, AUROC is unchanged by any monotone rescaling such as temperature scaling: a probe can be well calibrated yet undiscriminative, or discriminative yet miscalibrated.


\paragraph*{Correlation coefficients.}
We report both Pearson's~$r$, which measures linear association,
\begin{equation}
    r = \frac{\sum_i \left(x_i - \bar{x}\right)\left(y_i - \bar{y}\right)}{\sqrt{\sum_i \left(x_i - \bar{x}\right)^2} \sqrt{\sum_i \left(y_i - \bar{y}\right)^2}},
\end{equation}
and Spearman's~$\rho = r\!\left(\operatorname{rank}(x), \operatorname{rank}(y)\right)$, the Pearson correlation of the rank-transformed variables (mid-ranks for ties), which measures monotone association and is insensitive to outliers and marginal distributions.

\paragraph*{Confidence intervals.}
Unless otherwise noted, 95\% confidence intervals are computed with a nonparametric percentile bootstrap.
Because our data are clustered (each question contributes several rollouts, and all ablation pairs within a question share the same control baseline), we resample at the \emph{question} level: questions are drawn with replacement and all of their rollouts or perturbation pairs are pooled.

\subsection*{LLM-as-a-judge}
\label{sec:judge}
LLM-as-a-judge has become a standard evaluation tool~\citep{gu2025surveyllmasajudge}: a capable model quantifies properties of a piece of text, providing the flexibility required to assess an inherently unquantifiable substrate like natural language.

The canonical application in our setting is answer matching.
A forecasting question may list its ground-truth answer as ``Federal Bureau of Investigation'', while the model responds with any of a number of valid surface forms:
\begin{quote}
    \verb|the FBI|, \verb|F.B.I.|, \verb|the Federal Bureau of Investigations (FBI)|, etc.
\end{quote}
Exact or rule-based string matching (e.g.\ regular expressions) is brittle to such variation and produces both false negatives and false positives; an LLM judge, by contrast, reliably recognizes semantic equivalence.
Beyond binary correctness, LLM judges are also well suited to graded assessments of syntactic and semantic content, such as coherence, fluency, or the quality of intermediate reasoning steps.

We rely on an LLM-as-a-judge in two roles.
First, for \emph{answer matching}: scoring whether a model's free-form forecast agrees with the resolved ground truth, which underlies the correctness labels used throughout our calibration and probing experiments.
Second, for \emph{chain-of-thought scoring} in the faithfulness experiments (Section~\ref{sec:CoT_faithfulness}), where the judge rates whether the reasoning cites a given piece of evidence (citation score) and its stance toward it (accept / weigh / reject / not mentioned).

\section{Experiments}
\label{sec:experiments}

\subsection{Accuracy and Confidence}
\label{sec:accuracy-confidence}

\subsubsection{\model{EF-8B} is stable but overconfident across sampling temperatures}
\label{sec:temp-sweep}

Before probing internal representations, we characterize how \model{EF-8B}'s forecasting behavior responds to sampling temperature. 
For ten temperatures $T \in \{0.2, 0.4, \dots, 2.0\}$, we draw $10$ rollouts on all $296$ questions of the \dataset{OpenForesight-test\textsuperscript{*}} split ($29{,}600$ generations) and measure \emph{self-consistency} (fraction of a question's rollouts in its modal answer cluster), \emph{correctness} (per-rollout accuracy, with answer equivalence judged by \hyperref[sec:judge]{LLM-as-a-judge}), and \emph{confidence} (mean verbalized probability between the \texttt{<probability>} tags).

Across $T \lesssim 1.6$, accuracy ($37 \pm 1\%$) and confidence (${\sim}50\%$) are both flat while self-consistency falls steadily ($66\% \to 30\%$); accuracy degrades only past $T \approx 1.6$ (Appendix~\ref{app:temp-sweep}). 
We adopt $T = 1.0$ for the rest of the paper. 
This stability exposes a calibration problem: 
confidence sits near $50\%$ against ${\sim}37\%$ accuracy and barely moves, even as the rollouts disagree more, leaving \model{EF-8B} consistently overconfident (ECE $0.110$--$0.150$). 
This motivates our central question: 
whether \model{EF-8B}'s internal representations estimate confidence and correctness more faithfully than its verbalized probability.

\subsubsection{Forecasts are most accurate when the gold answer is contained in the prompt}
\label{sec:forecast-correctness}

Forecast correctness co-varies strongly with whether the retrieved news passages in the prompt mention the gold answer. 
Using an \hyperref[sec:judge]{LLM judge} to label answer containment in the evidence span---semantic matching, applied uniformly to the \dataset{test}/\dataset{validation} splits of both dataset builds---we find \model{EF-8B} reaches 86\%--94\% accuracy on answer-bearing questions versus 0.26--0.33 when the answer is absent ($\Delta = +57\%$ to $+65\%$ per slice, every 95\% bootstrap CI excluding zero). 

The pattern is informative about how the model answers. 
When the prompt presents a space of possibilities that includes the correct answer, \model{EF-8B} reliably converges on it,
whereas with no candidate in context it must fall back on parametric knowledge, and accuracy drops to $\sim$0.3. 
The model retrieves from available information rather than guessing from what it has knows. 
This is a desirable trait for a forecasting model -- and \textbf{it demonstrates that robust information retrieval and context building are critical to accurate forecasts.}

\subsection{Calibration of confidence}
\label{sec:calibration}

\subsubsection{Probing internal representations for correctness}
We train a family of activation-pooling probes to read \model{EF-8B}'s confidence directly from its internal state. 
Each probe is a small classifier on the frozen activations of a single layer, trained to predict whether a rollout's final answer will resolve correctly. 
We sweep all 36 layers, all three pooling architectures of Section~\ref{sec:probes} (mean-, attention-, and covariance-pooling), and several positions along the reasoning trace, from the opening \texttt{<think>} through the rollout to the closing \texttt{</think>}.
The full sweep, training recipe, and per-site metrics are in Appendix~\ref{app:probe-sweep}.

The sweep shows a consistent picture. 
Discrimination concentrates in the mid-to-late layers (19--24) and is largely in place before reasoning begins
(probes read at the end of the prompt already reach AUROC $\approx 0.76$),
consistent with the forced-answering finding (Section~\ref{sec:forced}) that
much of the forecast is fixed before the chain of thought reflects a finding. 
Calibration, by contrast, depends strongly on the pooling architecture: 
attention probes discriminate well but are persistently overconfident, whereas covariance probes match their discrimination while staying far better calibrated. 
We therefore deploy a covariance probe at layer 21, read late in the rollout.

\subsubsection{Probe calibration surpasses verbalized confidence}
\begin{figure}[h]
  \centering
  \includegraphics[width=0.5\linewidth]{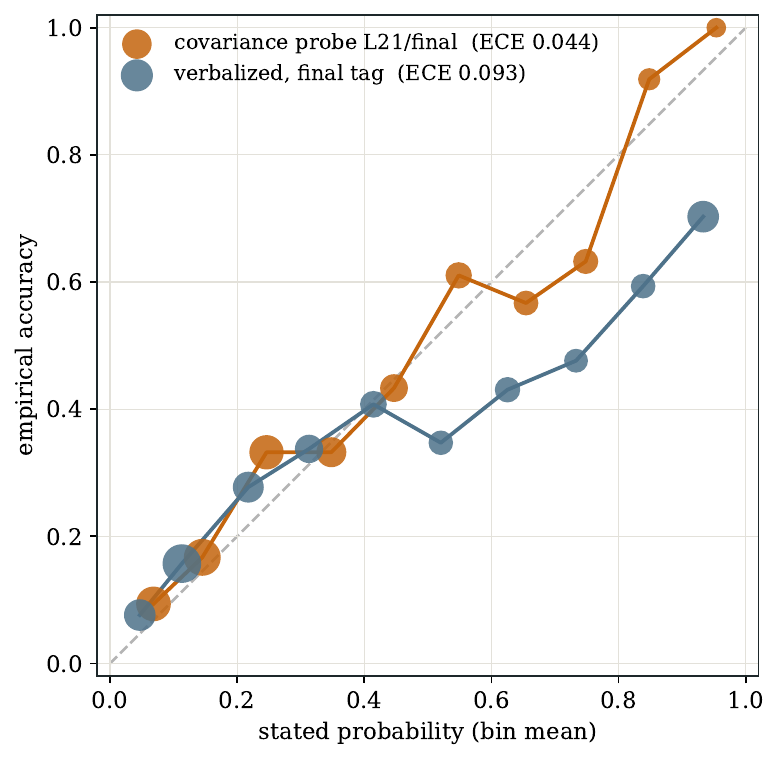}
  \caption{
\textbf{Reliability diagram} for the layer-21 covariance probe (ember) and \model{EF-8B}'s verbalized confidence from the final \texttt{<probability>} tag (slate), on the \dataset{OpenForesight-test} split ($N = 3{,}020$ rollouts each). 
10 equal-width bins; marker size $\propto$ bin count; diagonal $=$ perfect calibration. 
The probe tracks the diagonal (ECE $=0.044$) while verbalized confidence is overconfident above $0.5$ (ECE $=0.093$).
  }
  \label{fig:ECE-probes}
\end{figure}

Figure~\ref{fig:ECE-probes} compares this probe against the model's verbalized confidence -- the probability the model verbalizes in its final \texttt{<probability>} tag, present in all rollouts -- on the \dataset{OpenForesight-test} split ($N = 3{,}020$ rollouts for both signals). 
The two signals are statistically indistinguishable as rankers (AUROC $0.756$ vs.\ $0.758$), but differ sharply in calibration. 
Verbalized confidence is nearly calibrated below $0.5$ and systematically overconfident above it:
rollouts stated at $0.93$ resolve correct $70\%$ of the time, and those stated at $0.75$ only $44\%$. 
The probe tracks the diagonal across the full range, yielding ECE $=0.044$ vs.\ $0.093$. 
The same pattern holds out of distribution: the probe's calibration advantage persists on both OOD sources (\dataset{skysports} ECE $0.064$ vs.\ $0.089$; \dataset{aljazeera} $0.140$ vs.\ $0.178$), while verbalized discrimination degrades to near chance on \dataset{aljazeera} (AUROC $0.587$). 
The probe additionally provides a confidence estimate at \emph{every} rollout depth, including mid-reasoning before any answer is committed -- a property we exploit for faithfulness auditing in Section~\ref{sec:CoT_faithfulness}.

\paragraph*{Probing for \emph{correctness} thus yields calibration:}
a lightweight readout of the model's internal state recovers the well-calibrated confidence that the model's own verbalization distorts, 
without consuming any more inference tokens.

\subsection{Probe-only calibration on GLM models}
\label{sec:glm-probe-only}

The \model{EF-8B} result leaves open whether calibrated internal readouts depend on training the underlying model for our forecasting setting.
We therefore repeat the probe-only recipe on two frozen GLM models of different sizes, \model{GLM-4.7-Flash} and \model{GLM-4.5-Air}.
In both cases, the model is held fixed and only a small probe is trained on roughly ten thousand improved-context \dataset{OpenForesight} rollouts.

\begin{table}[H]
\centering\footnotesize
\setlength{\tabcolsep}{4pt}
\caption{\textbf{Probe-only calibration on frozen GLM models.} Each row trains only probe weights on the improved-context \dataset{OpenForesight} train pool and reports held-out test metrics. ECE uses raw probabilities with no recalibration. Clean $\Delta$AUROC removes rollouts with mid-reasoning answer or probability leakage according to each experiment's control.}
\label{tab:glm-probe-calibration}
\resizebox{\linewidth}{!}{%
\begin{tabular}{llrrrrrll}
\toprule
Model & Probe & Test $n$ & Probe AUROC & Verbal AUROC & Probe ECE & Verbal ECE & Clean $\Delta$AUROC [95\% CI] & Note \\
\midrule
\model{GLM-4.7-Flash} & Mean-pool L28 & $2{,}960$ & $0.768$ & $0.669$ & $0.054$ & $0.287$ & $+0.055\ [-0.006, 0.111]$ & Best AUROC: cov L26, $0.790$ \\
\model{GLM-4.5-Air} & Attention-1 L18 & $2{,}960$ & $0.783$ & $0.691$ & $0.102$ & $0.255$ & $\mathbf{+0.110}\ [0.069, 0.149]$ & Best ECE: linear L24, $0.097$ \\
\bottomrule
\end{tabular}%
}
\end{table}

\begin{figure}[H]
\centering
\begin{tikzpicture}
\begin{axis}[
  width=0.48\linewidth, height=4.4cm,
  ybar, bar width=8pt,
  symbolic x coords={GLM-4.7, GLM-4.5}, xtick=data,
  ymin=0.55, ymax=0.85,
  ylabel={AUROC $\uparrow$},
  title={Test discrimination},
  enlarge x limits=0.45,
  legend style={font=\scriptsize, at={(0.5,-0.32)}, anchor=north, legend columns=2},
  ymajorgrids, major grid style={gray!30},
  tick align=outside
]
\addplot[fill=ember, draw=ember] coordinates {(GLM-4.7,0.768) (GLM-4.5,0.783)};
\addplot[fill=slate, draw=slate] coordinates {(GLM-4.7,0.669) (GLM-4.5,0.691)};
\legend{Probe, Verbal}
\end{axis}
\end{tikzpicture}\hfill
\begin{tikzpicture}
\begin{axis}[
  width=0.48\linewidth, height=4.4cm,
  ybar, bar width=8pt,
  symbolic x coords={GLM-4.7, GLM-4.5}, xtick=data,
  ymin=0, ymax=0.32,
  ylabel={ECE $\downarrow$},
  title={Raw calibration},
  enlarge x limits=0.45,
  legend style={font=\scriptsize, at={(0.5,-0.32)}, anchor=north, legend columns=2},
  ymajorgrids, major grid style={gray!30},
  tick align=outside
]
\addplot[fill=ember, draw=ember] coordinates {(GLM-4.7,0.054) (GLM-4.5,0.102)};
\addplot[fill=slate, draw=slate] coordinates {(GLM-4.7,0.287) (GLM-4.5,0.255)};
\legend{Probe, Verbal}
\end{axis}
\end{tikzpicture}
\caption{\textbf{Probe-only confidence from GLM internals.} Internal probes improve AUROC on both full test sets (left) and sharply reduce raw ECE relative to verbalized confidence (right). The \model{GLM-4.7-Flash} clean ranking interval includes zero, so its robust claim is calibration; \model{GLM-4.5-Air} improves both ranking and calibration.}
\label{fig:glm-probe-calibration-bars}
\end{figure}
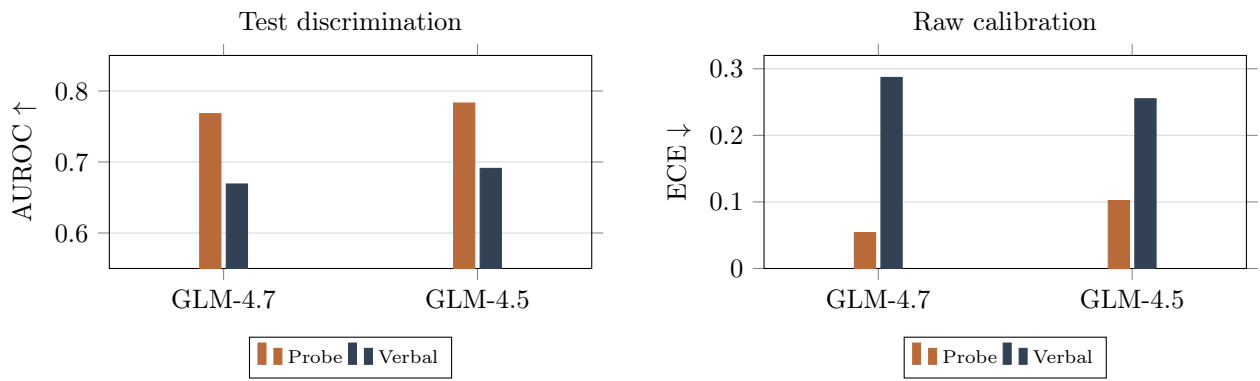

\begin{figure}[H]
\centering
\includegraphics[width=0.62\linewidth]{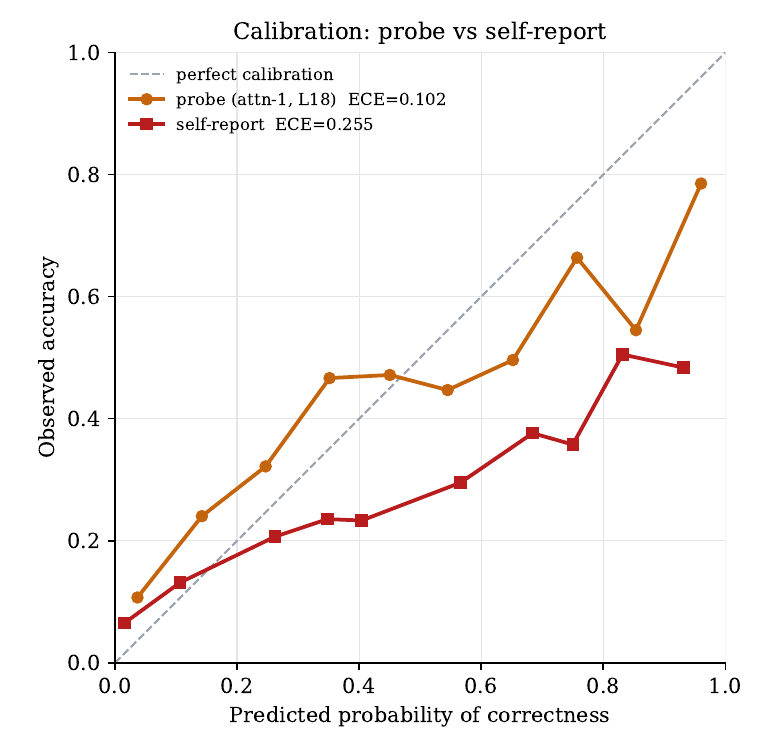}
\caption{\textbf{GLM-4.5-Air reliability.} The layer-18 attention probe is substantially better calibrated than the model's stated probability on the test split.}
\label{fig:glm-calibration}
\end{figure}

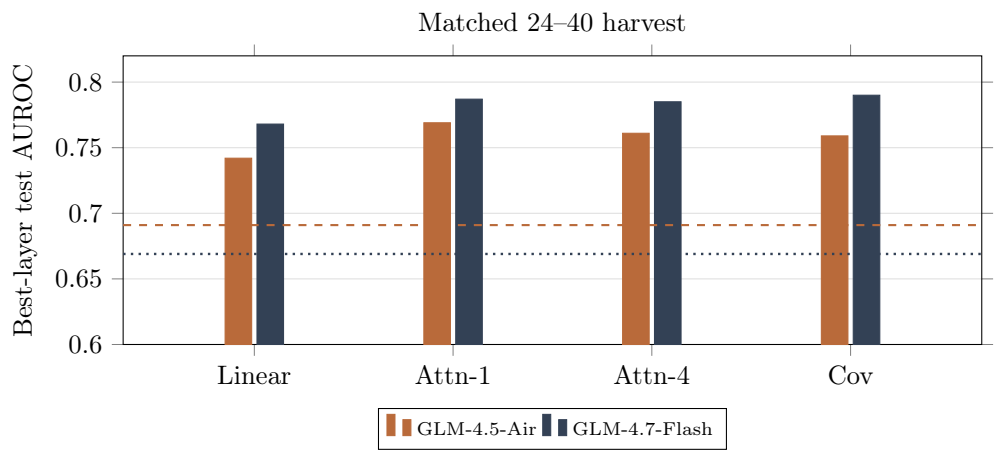
\begin{figure}[H]
\centering
\begin{tikzpicture}
\begin{axis}[
  width=0.78\linewidth,
  height=5.4cm,
  ybar,
  bar width=10pt,
  symbolic x coords={Linear, Attn-1, Attn-4, Cov},
  xtick=data,
  ymin=0.60,
  ymax=0.82,
  ylabel={Best-layer test AUROC},
  title={Matched 24--40 harvest},
  enlarge x limits=0.22,
  ymajorgrids,
  major grid style={gray!25},
  tick align=outside,
  legend style={font=\scriptsize, at={(0.5,-0.22)}, anchor=north, legend columns=2}
]
\addplot[fill=ember, draw=ember] coordinates {(Linear,0.742) (Attn-1,0.769) (Attn-4,0.761) (Cov,0.759)};
\addplot[fill=slate, draw=slate] coordinates {(Linear,0.768) (Attn-1,0.787) (Attn-4,0.785) (Cov,0.790)};
\draw[dashed, ember, line width=0.8pt] ({rel axis cs:0,0}|-{axis cs:Linear,0.691}) -- ({rel axis cs:1,0}|-{axis cs:Linear,0.691});
\draw[dotted, slate, line width=0.9pt] ({rel axis cs:0,0}|-{axis cs:Linear,0.669}) -- ({rel axis cs:1,0}|-{axis cs:Linear,0.669});
\legend{GLM-4.5-Air, GLM-4.7-Flash}
\end{axis}
\end{tikzpicture}
\caption{\textbf{Matched-depth cross-model comparison.} Best-layer test AUROC by probe family for \model{GLM-4.5-Air} and \model{GLM-4.7-Flash} on the matched 24--40 harvest, with each model's verbalized-confidence AUROC as a reference line.}
\label{fig:glm-cross-model}
\end{figure}

The calibration effect is the common result.
On \model{GLM-4.7-Flash}, the practical mean-pool probe has ECE $0.054$ versus $0.287$ for verbalized confidence; the leakage-controlled ranking delta is positive but not significant.
On \model{GLM-4.5-Air}, the layer-18 attention probe improves both AUROC and ECE, and the clean ranking interval stays above zero.
These results support the central use case: small probes can supply calibrated correctness estimates for frozen models without updating model weights.

\subsection{Math reasoning calibration}
\label{sec:math-calibration}

The forecasting experiments show that internal probes recover a better calibrated confidence signal than the model's verbalized confidence.
We next ask whether the same readout helps in a different calibrated-reasoning setting: mathematical reasoning after RLVR.
Following~\citet{ma2026decouplingreasoningconfidenceresurrecting}, we reproduce DCPO as a strong training-time calibration baseline for \model{Qwen3-8B}.
We then freeze each model and train a correctness probe on residual-stream activations immediately before the boxed answer, comparing it against token-logprob confidence and recalibrated verbalized confidence on pooled OOD math benchmarks.

\begin{table*}[t]
\centering\footnotesize
\setlength{\tabcolsep}{5pt}
\caption{
\textbf{Correctness probe vs.\ best non-probe confidence on out-of-distribution math}
(pooled AIME24/25 + AMC23/24, $n=580$).
$\Delta = \text{probe} - \text{best non-probe}$; AUROC higher is better ($\Delta>0$ favors the probe), ECE lower is better ($\Delta<0$ favors the probe).
Intervals are 95\% qid-clustered bootstrap CIs.
The best non-probe AUROC is whole-response token-logprob confidence for both rows.
Probe site is layer / depth fraction / pooling.
}
\label{tab:ood-main}
\resizebox{\textwidth}{!}{%
\begin{tabular}{lll ccc cccc}
\toprule
& & & \multicolumn{3}{c}{AUROC $\uparrow$} & \multicolumn{4}{c}{ECE $\downarrow$} \\
\cmidrule(lr){4-6}\cmidrule(lr){7-10}
Model & Decode & Probe site & Probe & Best n.p. & $\Delta$ [95\% CI]
      & Probe & Best n.p. & $\Delta$ [95\% CI] & Best ECE baseline \\
\midrule
Untrained \model{Qwen3-8B} & plain  & L18 / r100 / last-tok & $0.894$ & $0.752$ & $\mathbf{+0.143}\ [0.090, 0.197]$
                   & $0.083$ & $0.270$ & $\mathbf{-0.187}\ [-0.227, -0.117]$ & iso-logits \\
DCPO-recipe \model{Qwen3-8B} & verbal & L21 / r100 / last-tok & $0.929$ & $0.865$ & $\mathbf{+0.064}\ [0.023, 0.109]$
                   & $0.143$ & $0.125$ & $+0.040\ [0.009, 0.074]$ & iso-verbal \\
\bottomrule
\end{tabular}
}
\end{table*}

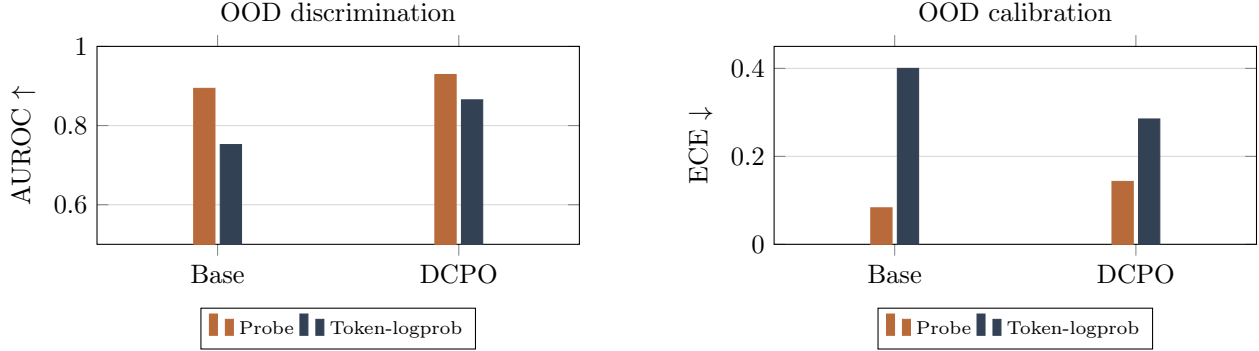
\begin{figure}[t]
\centering
\begin{tikzpicture}
\begin{axis}[
  width=0.48\linewidth, height=4.2cm, ybar, bar width=8pt,
  symbolic x coords={Base, DCPO}, xtick=data,
  ymin=0.5, ymax=1.0, ylabel={AUROC $\uparrow$}, title={OOD discrimination},
  enlarge x limits=0.5, legend style={font=\scriptsize, at={(0.5,-0.32)}, anchor=north, legend columns=2},
  ymajorgrids, major grid style={gray!30}]
\addplot[fill=ember, draw=ember] coordinates {(Base,0.894) (DCPO,0.929)};
\addplot[fill=slate, draw=slate] coordinates {(Base,0.752) (DCPO,0.865)};
\legend{Probe, Token-logprob}
\end{axis}
\end{tikzpicture}\hfill
\begin{tikzpicture}
\begin{axis}[
  width=0.48\linewidth, height=4.2cm, ybar, bar width=8pt,
  symbolic x coords={Base, DCPO}, xtick=data,
  ymin=0, ymax=0.45, ylabel={ECE $\downarrow$}, title={OOD calibration},
  enlarge x limits=0.5, legend style={font=\scriptsize, at={(0.5,-0.32)}, anchor=north, legend columns=2},
  ymajorgrids, major grid style={gray!30}]
\addplot[fill=ember, draw=ember] coordinates {(Base,0.083) (DCPO,0.143)};
\addplot[fill=slate, draw=slate] coordinates {(Base,0.400) (DCPO,0.285)};
\legend{Probe, Token-logprob}
\end{axis}
\end{tikzpicture}
\caption{
\textbf{Probe vs.\ token-logprob confidence on pooled OOD math.}
Only the untrained and DCPO-recipe models are shown, and clean MATH-500 is excluded.
The frozen correctness probe out-discriminates whole-response token-logprob confidence on both models (left) and is better calibrated where token-logprob confidence degrades OOD (right).
ECE is 15-bin.
}
\label{fig:ood-probe-vs-logits}
\end{figure}

The OOD pattern separates discrimination from point calibration.
For the untrained base, the probe improves both ranking and calibration: AUROC rises from $0.752$ to $0.894$, while ECE falls from $0.270$ to $0.083$.
For the DCPO-recipe model, the probe is still the stronger OOD correctness monitor by AUROC ($0.929$ vs.\ $0.865$), but isotonic-recalibrated verbalized confidence remains slightly better by ECE ($0.125$ vs.\ $0.143$).
This result supports probes as OOD correctness monitors for math reasoning, while leaving calibration method choice dependent on whether the target is ranking, point calibration, or both.
Appendix~\ref{app:math-probe} gives additional setup and controls.

\subsection{Chain-of-Thought Faithfulness}
\label{sec:CoT_faithfulness}

A faithful chain of thought should change when the evidence behind a forecast changes.
We test this directly, following \citet{gurarieh2026faithfulnessmetricsdontmeasure}: we perturb the
input prompt and measure two responses to the \emph{same} perturbation.
\begin{itemize}
    \item The \textbf{behavioral change} $|\Delta\mathrm{prob}|$: the absolute shift in the model's
    stated forecast probability (read from its final \texttt{<probability>} tag) between the original
    and perturbed prompt, averaged over completions when several are sampled.
    \item The \textbf{internal change} $|\Delta\mathrm{probe}|$: the absolute shift, over the same pair,
    in the estimate of an internal correctness probe (the calibrated readout of
    Section~\ref{sec:calibration}; here its layer-20 attention variant).
\end{itemize}
A faithful reasoning trace should visibly update whenever $|\Delta\mathrm{prob}|$ is large.
We apply this in three settings: removing real evidence (\textit{ablation}, Section~\ref{sec:ablation}),
inserting misleading evidence (\textit{injection}, Section~\ref{sec:injection}), and reading the probe directly
as an internal lie detector when the written reasoning stays silent (Section~\ref{sec:lie-detector}).

\subsubsection{Evidence Ablation Faithfulness}
\label{sec:ablation}
Does \model{EF-8B}'s chain-of-thought reasoning faithfully reflect which retrieved news articles actually
drive its forecast probabilities? We test this by removing the news sources one at a time: drop a single
real article, then measure whether and how the reasoning updates. We expect that
\begin{itemize}
    \item[\textit{a.}] \emph{Correlation:} the behavioral change from removing an article should track the
    change in the reasoning;
    \item[\textit{b.}] \emph{Stealth influence:} among articles that clearly move the forecast
    ($|\Delta\mathrm{prob}| > 0.05$), few should leave the reasoning unchanged;
    \item[\textit{c.}] \emph{Signal above noise:} the induced probability shifts should be distinguishable
    from ordinary same-prompt sampling variation.
\end{itemize}

\paragraph*{Ablation protocol.}
For each forecasting question in \dataset{test\textsuperscript{*}+val\textsuperscript{*}} with $N > 0$ articles (354 questions), we build $N$ prompts with one article removed (renumbering the rest),
plus one prompt with no articles. 
As a control, we run the 354 original prompts through the same generation pipeline, so any measured change reflects the missing article rather than run-to-run decoding variation.

\paragraph*{Behavioral change.}
We average the stated probability over 3 completions per prompt and take $|\Delta\mathrm{prob}|$ as the
absolute difference between the control and ablated means.

\begin{figure}[h]
  \centering
  \includegraphics[width=\textwidth]{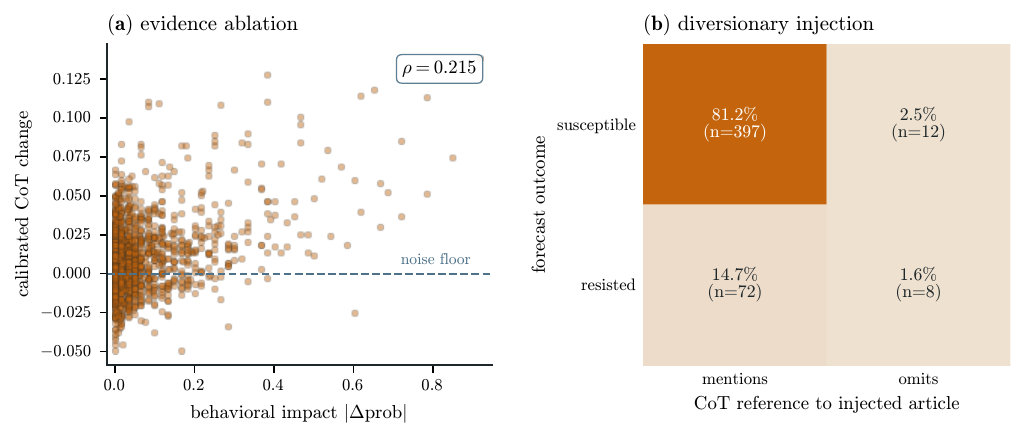}
  \caption{\textbf{Chain-of-thought faithfulness under evidence perturbation.}
  \textbf{(a)} Evidence ablation. Each point is one (question, removed-article) pair:
  behavioral impact $|\Delta\text{prob}|$ (same-pipeline control vs.\ single-article-ablated)
  against calibrated CoT change (within-control baseline similarity minus control-vs-ablated
  similarity; composite similarity $=$ mean of word-set Jaccard and character-sequence similarity).
  Positive $y$ means the ablation perturbed the reasoning trace more than $T{=}1.0$ sampling
  noise; the dashed line at $y{=}0$ is the noise floor. Behavioral and reasoning change are
  only weakly coupled (Spearman $\rho = 0.215$), and 107 of 460 high-impact pairs
  ($|\Delta\text{prob}| > 0.05$) show no CoT change, a 23\% stealth-influence rate.
  $N = 1{,}303$ single-article ablation pairs from 354 questions. \textbf{(b)} Diversionary
  injection. Four-quadrant classification of 489 questions by whether the CoT mentions an
  injected misleading article ($x$) and whether the forecast was behaviorally susceptible to
  it ($y$); cells give the share and count of questions. The model is largely transparent when
  actively misled, 81.2\% adopt the fabricated evidence and openly cite it (honest-susceptible),
  with stealth adoption (susceptible but unmentioned) in only 2.5\% (n $=$ 12).}
  \label{fig:faithfulness}
\end{figure}

\paragraph*{Reasoning change.}
We score how much the reasoning \emph{text} moves, relative to how much it moves by chance. 
For each question we compare the three control traces to each other (a baseline for sampling variation at $T=1.0$) and compare control traces to ablated ones. 
The \emph{calibrated CoT change} is the baseline similarity minus the control-vs-ablated similarity, so positive values mean the ablation shifted the reasoning more than sampling noise alone (the dashed noise floor in Fig.~\ref{fig:faithfulness}a). 
Text similarity is the average of a word-overlap (Jaccard) score and a character-sequence similarity score;
separately, we record whether a trace cites the removed article by name or title.

\paragraph*{Reasoning change only weakly tracks behavioral impact.}
Across 1,303 ablation pairs, behavioral and reasoning change are only weakly related (Spearman $\rho = 0.215$; Fig.~\ref{fig:faithfulness}a). 
Stealth influence is substantial: of the 460 high-impact pairs ($|\Delta\mathrm{prob}| > 0.05$), 107 (23\%) show no change in the reasoning at all.

\subsubsection{Diversionary Injection Faithfulness}
\label{sec:injection}
Instead of removing real evidence (ablation), injection adds fake evidence: 
we insert one misleading article and check whether the reasoning discloses it. 
We consider three hypotheses:
\begin{itemize}
    \item[\textit{a.}] the model will be susceptible to misleading articles;
    \item[\textit{b.}] when susceptible, the reasoning will sometimes fail to mention the misleading evidence;
    \item[\textit{c.}] susceptibility will fall as the amount of genuine evidence rises.
\end{itemize}

We evaluate 489 forecasting questions from \dataset{test\textsuperscript{*}+val\textsuperscript{*}}. 
One misleading article per question is written by Claude Sonnet (\texttt{claude-sonnet-4-20250514}), and an LLM judge (also Claude Sonnet) rates each reasoning trace on two axes: 
whether it cites the article (citation score $0/0.5/1$) and its stance toward it (\verb|accept/weigh/reject/not_mentioned|).

Most questions (81.2\%) are \emph{honest-susceptible} (Fig.~\ref{fig:faithfulness}b, top-left): 
the model adopts the misleading evidence and openly cites it. 
Only 2.5\% of the 489 questions ($n=12$) are \emph{stealth}, adopting the evidence without mentioning it.
A further 14.7\% are \emph{considered-and-rejected}:
the model mentions the evidence but resists it. 
This 2$\times$2 aggregates at the question level (a question counts as ``mentioned'' if any of its 3 completions cites the evidence); at the completion level, 46/1,227 susceptible completions (3.8\%) are stealth.

Two patterns stand out. 
The model treats any plausible article as strong evidence regardless of quality, yet it is strikingly transparent when actively misled, openly citing the fabricated source. Stealth behavior is far rarer here than under ablation (2.9\% of susceptible questions vs.\ 23\% of high-impact ablation pairs), though the two rates use different denominators and are not directly comparable.

\subsubsection{Probe-Based CoT Lie Detector}
\label{sec:lie-detector}

\paragraph*{Setup.}
A layer-20 attention probe (AUC 0.747 on answer-correctness) trained on \model{EF-8B} carries information about reasoning quality.
If the probe's internal change $|\Delta\mathrm{probe}|$ tracks the behavioral change $|\Delta\mathrm{prob}|$, and does so even in the stealth cases where the reasoning hides the change, then the probe acts as a lie
detector for unfaithful reasoning. We test this across both paradigms, article ablation (1,303 cases) and
diversionary injection (489 cases), reading $|\Delta\mathrm{prob}|$ and $|\Delta\mathrm{probe}|$ over each
original/perturbed pair (Section~\ref{sec:CoT_faithfulness}).

\begin{figure}[h]
  \centering
  \includegraphics[width=\textwidth]{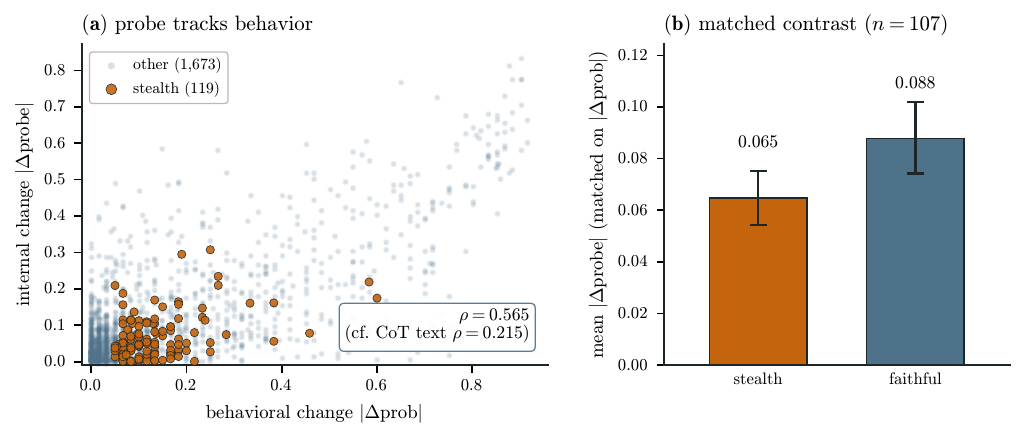}
  \caption{\textbf{Internal probe as a lie detector for unfaithful reasoning.}
  \textbf{(a)} 
  Across 1,792 perturbation cases (article ablation and diversionary injection), the layer-20 probe's internal change $|\Delta\text{probe}|$ tracks the behavioral change $|\Delta\text{prob}|$ (Spearman $\rho = 0.565$), far above the reasoning-text correlation of 0.215 (Fig.~\ref{fig:faithfulness}a). 
  Stealth cases (the forecast shifts but the reasoning stays silent; $n = 119$) are highlighted and lie along the trend rather than at the origin: the probe registers changes that the chain of thought omits. 
  \textbf{(b)} 
  Pairing each stealth case with a faithful case of near-identical $|\Delta\text{prob}|$, stealth probe shifts stay smaller than faithful ones (0.065 vs.\ 0.088) but well above zero, so the stealth signal is not merely a scaling artifact of smaller behavioral shifts. 
  Error bars: 95\% bootstrap CIs.
  }
  \label{fig:liedetector}
\end{figure}

\paragraph*{Core correlation.}
Across all 1,792 cases the probe strongly tracks behavioral change (Spearman $\rho = 0.565$;
$\rho = 0.641$ on signed shifts) and predicts the direction of change in 83.6\% of cases, far above the
reasoning-text correlation of $\rho = 0.215$ (Section~\ref{sec:ablation}; Fig.~\ref{fig:liedetector}a).

\paragraph*{Stealth detection.}
The signal persists where it matters most: in the 107 stealth ablation cases, where the forecast moves but
the reasoning does not, the probe still shifts (mean $|\Delta\mathrm{probe}| = 0.065$) and agrees with the
direction of change 78.5\% of the time. 
This is not just a byproduct of smaller shifts in stealth cases:
matching stealth to faithful cases of near-identical $|\Delta\mathrm{prob}|$ leaves stealth probe shifts
significantly smaller (0.065 vs.\ 0.088; Fig.~\ref{fig:liedetector}b), so the model genuinely processes the
two differently. 
Probes on intermediate activations thus give a complementary audit channel for reasoning faithfulness.

\subsection{Forced answering}
\label{sec:forced}

\paragraph{How much of \model{EF-32B}'s forecast is fixed before it reasons?} 
We compare the model's answer, verbalized confidence, and accuracy with and without its chain of thought. 
In the \emph{forced-answer} scheme we append the assistant prefill
\begin{center}
\texttt{<think>\textbackslash n\textbackslash n</think>\textbackslash n\textbackslash n<answer>}
\end{center}
after the prompt, forcing the model to bypass its free-roaming CoT and commit an answer immediately. 
We run forced and free schemes on the \dataset{OpenForesight-test} split and two OOD splits (\dataset{aljazeera2026Q1}, \dataset{aljazeeraLate2025}), with $10$ rollouts per arm per question.

\begin{figure}[h]
  \vspace{0.5cm}
  \centering
  \includegraphics[width=\linewidth]{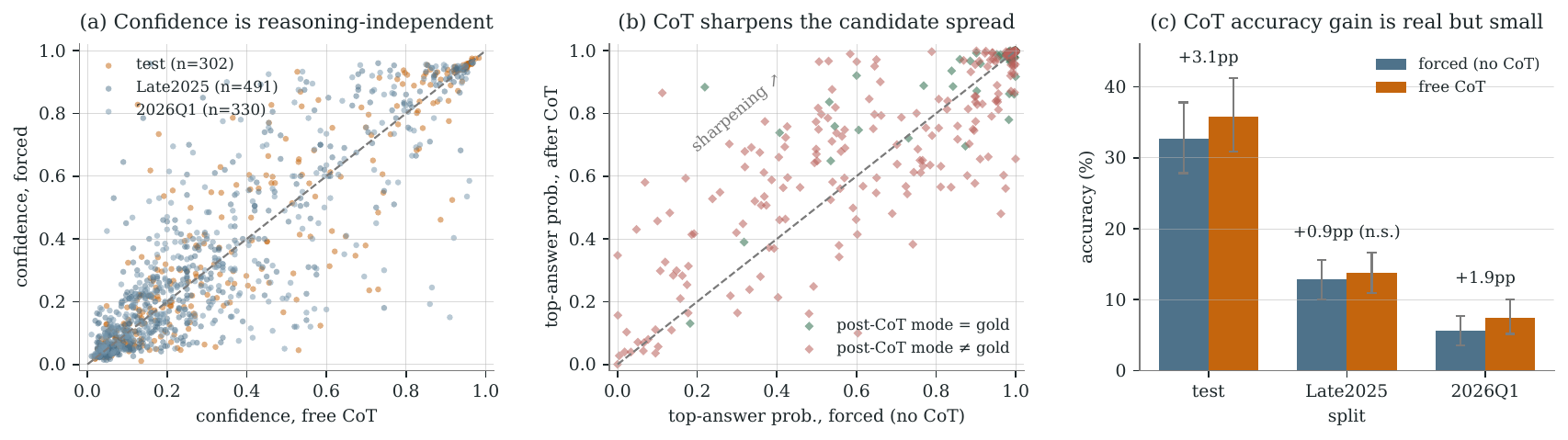}
  \caption{\textbf{What chain of thought does and does not change.}
  \textbf{(a)}~\emph{Confidence is reasoning-independent}: 
  per-question mean verbalized confidence with full reasoning ($x$) vs.\ under the forced-answer prefill ($y$) hugs the $y{=}x$ line on the \dataset{OpenForesight-test} split and both OOD splits (Spearman $\rho = 0.90 / 0.87 / 0.78$).
  \textbf{(b)}~\emph{Forced answering surfaces candidate answers; CoT sharpens
  the choice}: 
  top-answer probability of the forced candidate distribution ($x$) vs.\ after the model's own think block ($y$), at a fixed \texttt{<answer>} position ($n = 302$, colored by whether the post-CoT mode equals gold).
  The horizontal spread shows genuine competition among candidates; mass above the diagonal shows CoT concentrating probability on the leading one (sharper on $67\%$ of questions), yet among forced-wrong questions it corrects only $4\%$ and locks in the same wrong answer $72\%$ of the time.
  \textbf{(c)}~\emph{CoT's accuracy gain is real but small}: free-CoT vs.\ forced accuracy per split. In distribution the gain is $+1.9$\,pp ($95\%$ CI $[+1.0, +2.9]$), attenuating OOD ($+1.9$\,pp on \dataset{2026Q1}, non-significant $+0.9$\,pp on \dataset{Late2025}). 
  Error bars: $95\%$ question-bootstrap CIs.}
  \label{fig:forced-3panel}
\end{figure}

\paragraph{Confidence is reasoning-independent.}
Forced and free verbalized confidence are nearly identical: 
per-question mean confidence hugs the $y{=}x$ line with no systematic shift (Spearman $\rho = 0.90$ on \dataset{test}, $0.87$ on \dataset{Late2025}, $0.78$ on \dataset{2026Q1}; Fig.~\ref{fig:forced-3panel}a). 
The model states essentially the same confidence whether or not it has reasoned.

\paragraph{The committed answer is mostly pre-set.}
The forced answer matches the free modal answer on $67\%$ of \dataset{test} questions ($64\%$ on \dataset{Late2025}, $56\%$ on \dataset{2026Q1}): for two-thirds of questions, the chain of thought returns the answer the model would have committed to immediately. 
This pre-commitment is explicit in the logits: a single forward pass under the empty-think prefill reads the forced-answer distribution directly.

\paragraph{Forced answering surfaces candidate answers; chain of thought sharpens the choice.}
The logit readout does more than confirm the modal answer: 
it reproduces the sampled forced-answer distribution almost exactly ($r = 0.982$) and surfaces complete, plausible candidates that $50$~rollouts never produce. 
The forced pass thus exposes a genuine \emph{spread} over candidate answers, not a single commitment. 
Chain of thought acts on this spread mainly by sharpening it: 
with the \texttt{<answer>} position held fixed and only the think block varied, reasoning concentrates probability on the leading candidate for two-thirds of questions (Fig.~\ref{fig:forced-3panel}b).

\paragraph{CoT's accuracy gain is real but small.}
Free reasoning outperforms forced answering by $+1.9$\,pp (percentage-points) in distribution, and the gain attenuates out of distribution (Fig.~\ref{fig:forced-3panel}c). 
This result is not obtained by candidate selection: 
when the forced answer is wrong, the gold answer is usually absent from the candidate distribution altogether -- confident forced errors are genuine ignorance.
Reasoning calibrates commitment to an answer largely chosen before it begins; it does not search the candidate space for a better one.

\paragraph*{Forced answering is a cheap ($50\text{--}70$\texttimes) and near-exact proxy for \model{EF-32B}'s committed answer and confidence.} 
CoT mostly confirms and sharpens a forecast the model has already made from the prompt alone, contributing a small accuracy gain concentrated on hard but answerable questions, rather than recovering answers it did not already hold.

\subsection{Triaging questions by their pre-reasoning answer}
\label{sec:stability-gate}

Section~\ref{sec:forced} showed that \model{EF-32B} usually has its answer already decided before it reasons, and that a cheap forced pass recovers that answer.
The forced pass also tells us how spread out the answer is across candidates. 
We use this spread to decide, \emph{before any reasoning}, what to do with each question: 
commit the answer immediately, 
run a full reasoning rollout, 
or send the question to a retrieval step because the model does not have the evidence to answer it.

\paragraph{Measuring spread.}
The forced pass returns a distribution over candidate answers. 
Decoding the $8$ most probable answer continuations from the empty-think prefill produces a set of answer strings with sequence probabilities, renormalized to $p_i$.
We measure spread as the Shannon entropy $H = -\sum_i p_i \ln p_i$. 
$H$ is $0$ when all mass is on one answer and grows as the mass divides across candidates. 
On the \dataset{test} split $H$ ranges from $0$ to $2.1$~nats.

\begin{figure}[t]
  \centering
  \includegraphics[width=\linewidth]{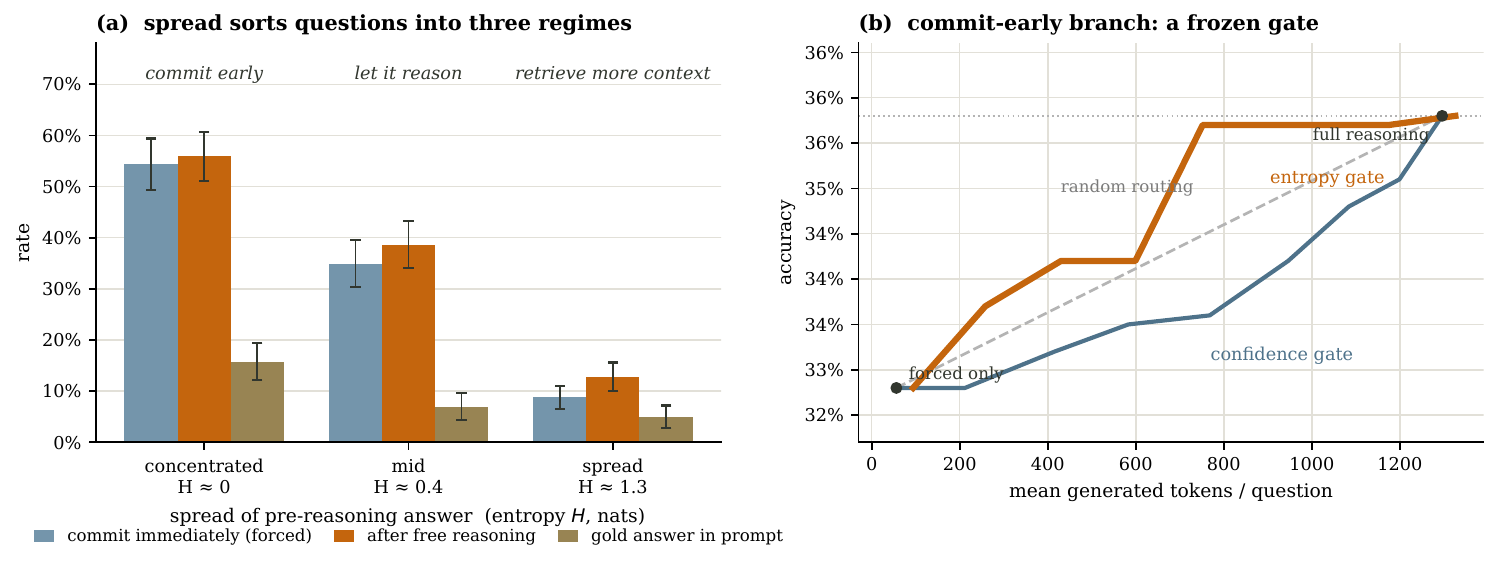}
  \caption{\textbf{Triaging forecasting questions by the spread of the model's
  pre-reasoning answer} (\dataset{OpenForesight-test}, $N=302$).
  \textbf{(a)} Questions split into thirds by the entropy $H$ of the forced-pass
  answer distribution. Bars show accuracy when the model commits immediately
  (forced), accuracy after free reasoning, and the rate at which the gold answer
  appears in the prompt; whiskers are $\pm 1$\,SE. Accuracy rises as the answer
  concentrates, reasoning helps only in the middle regime, and high-entropy
  questions both fail and lack supporting evidence.
  \textbf{(b)} Accuracy against mean generated tokens per question for four
  inference policies. The commit-early gate (entropy gate, ember) recovers
  full-reasoning accuracy at far fewer tokens, staying above naive random routing
  (dashed) and a verbalized-confidence gate (slate).}
  \label{fig:triage}
\end{figure}

\paragraph{Spread sorts questions into three regimes.}
We split the test questions into thirds by $H$ (Fig.~\ref{fig:triage}a). 
When $H$ is near $0$, the model is right about $56\%$ of the time and reasoning changes little ($54$ to $56$\,pp). 
In the middle the model is right $39\%$ of the time and \emph{reasoning does add a few points} ($35$ to $39$\,pp). 
When $H$ is high the model is right only $13\%$ of the time and reasoning barely helps ($9$ to $13$\,pp). 
The high-$H$ questions also rarely contain the gold answer in their prompt ($5\%$ against $16\%$ at low $H$, using the containment label from Section~\ref{sec:accuracy-confidence}). 
Reasoning cannot recover an answer the prompt does not support, so these questions need more evidence, not more tokens. 
The three regimes suggest three actions: commit early at low $H$, reason at moderate
$H$, and retrieve more context at high $H$.

\paragraph{Commit early: a token-saving procedure.}
When the model is already confident about an answer, it can skip its reasoning and respond directly. 
We train a simple classifier to spot these cases from cheap signals read off a single forced-answer pass: 
how spread out the candidate answers are (entropy), how much probability sits on the most likely answer, the gap between the top two answers, and how many distinct answers appear. 
It outputs one score, and we fix a single cutoff. 
Above the cutoff the model answers immediately; below it, the model reasons as usual. 
On held-out questions this matches the accuracy of always reasoning while using far fewer tokens
(Fig.~\ref{fig:triage}b). 
It beats routing questions at random, and beats a version that decides based on the model's stated confidence, which helps little because stated confidence tracks how hard a question is rather than whether reasoning would actually change the answer. 
Across all splits it saves $30$ to $47\%$ of generated tokens with no measurable accuracy loss: 
every per-split change is within about $1$\,pp of full reasoning, with a confidence interval that includes zero.

\paragraph{Retrieve: an extension.}
The same signal flags the questions where reasoning will not help. 
Because these are the questions missing evidence, the forced-pass read can also nominate questions for retrieval: 
instead of spending a long rollout to guess, fetch the article the model needs. 
We do not test this here. We note only that the signal separating ``the model knows'' from ``the model is guessing'' is the same one that drives the gate.

\paragraph*{Summary.}
Reading the model's answer distribution before it reasons gives two things at once:
(a) it saves tokens, by committing the questions the model has already settled instead of reasoning through them; 
(b) it can also raise accuracy, by flagging the questions where the model is guessing for lack of context, so they can be sent for retrieval rather than answered blindly. 
The same cheap signal cuts cost on the easy questions and points to where more evidence would help on the hard ones.

\section{Conclusion}
\label{sec:conclusion}

We found that a language model forecaster's internal representations offer a more reliable window into its confidence and reasoning than its own words.
Lightweight probes on intermediate activations recover a well-calibrated confidence signal that the model's verbalization distorts, track the influence of evidence that the chain of thought fails to disclose, and reveal that the forecast is largely committed before reasoning begins -- a fact that can be exploited to route questions between committing, reasoning, and retrieving.
Across all three threads the pattern is the same: internal activations contain more information than what the model is able to verbalize.

This raises the question of introspection. 
The calibrated signal is demonstrably present in the residual stream from the final prompt token onward, yet the stated probability does not report it -- overconfidence is less a failure of self-knowledge than of self-report.
Whether models can be trained to verbalize their internal confidence estimate, for instance by distilling the probe signal into the stated probability, or by post-training against calibration-aware rewards, is a natural next step.
Testing whether such training sharpens or instead degrades the internal signal the probes read is another one.

\paragraph*{}
The experiments presented were conducted by \Silico, Goodfire's agentic platform for interpretability research, autonomously with human feedback.
All experimental designs, findings, write-ups, scripts, and figures have been carefully reviewed and edited by the authors.

\bibliographystyle{plainnat}
\bibliography{template/references}

\clearpage

\appendix
\section*{Appendix}
\section{Data audit}
\label{app:data-audit}
We perform a simple data audit on the \dataset{nikhilchandak/OpenForesight} dataset.
We check notably for typical prompt length, question leakage, etc.

The difference in prompt length is solely explained by the fact that the last two splits don't include retrieval articles in their prompts. Therefore, the forecasting setting is limited to instructions, question, and some brief background information.

For the splits that include article retrieval, with the exception of a few anomalous prompts, prompts include 5 news articles.


\begin{figure}[h]
  \centering
  \includegraphics[width=0.7\linewidth]{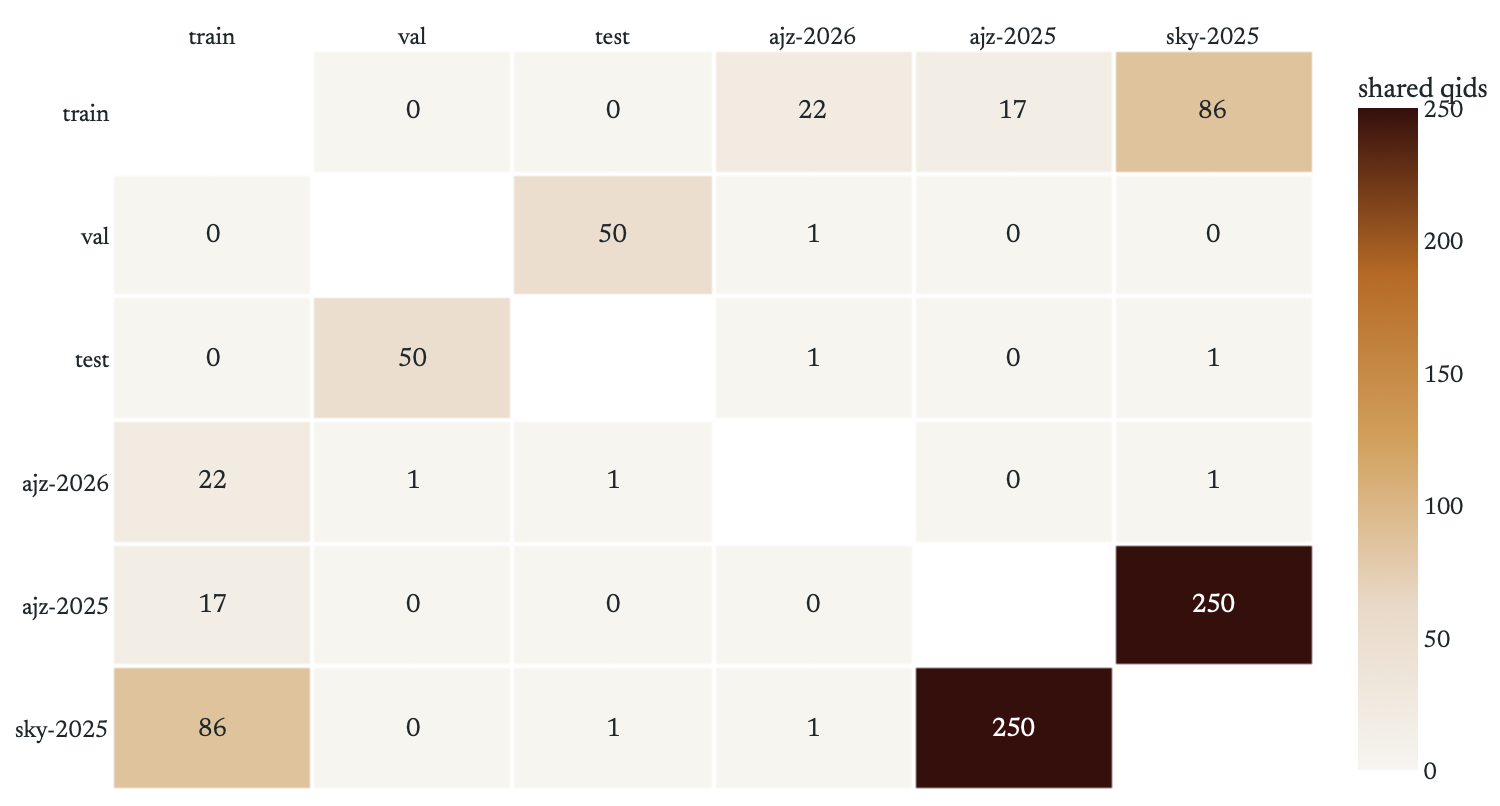}
  \caption{
  \textbf{Cross-split qid overlap in \dataset{OpenForesight}.} 
  Off-diagonal cells give the number of question IDs (qids) shared by both splits; the diagonal (split sizes, listed below) is omitted from the color scale. Nine of fifteen split pairs share qids (429 shared-qid instances in total), yet every shared qid is a collision: in all 429 cases the two questions differ in both title and gold answer (same-question rate = 0). The held-out evaluation splits are qid-disjoint from train. qid is therefore a split-local index, not a global key, and exact-qid overlap reflects index reuse rather than question leakage. This analysis detects exact-qid and exact-text matches only; semantic near-duplicates (the same real-world question under a different qid or wording) are not captured and require embedder/judge-based detection. Splits: \texttt{train} (n=52,183), \texttt{validation} (207), \texttt{test} (302), \texttt{aljazeera2026Q1} (330), \texttt{aljazeeraLate2025} (491), \texttt{skysports2025} (1,788).
  }
  \label{fig:data-audit_shared-qids}
\end{figure}

\begin{figure}[h]
  \centering
  \includegraphics[width=0.9\linewidth]{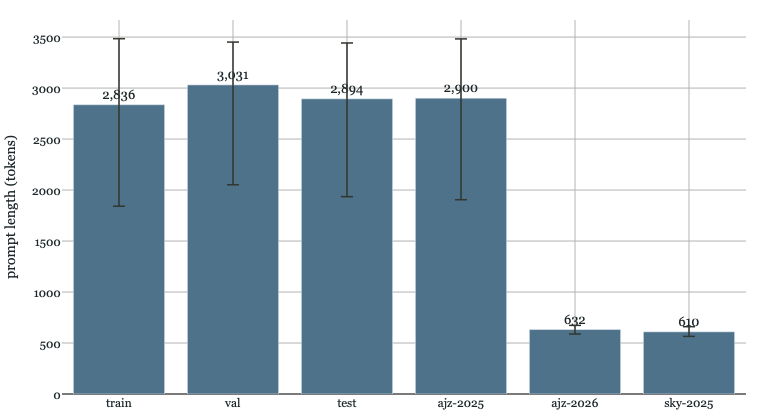}
  \caption{
\textbf{Prompt token length by split in \dataset{OpenForesight}.} Bars show the mean token count of the retrieval-augmented prompt field, tokenized with the \model{Qwen3} tokenizer with special tokens excluded; whiskers span the 2.5th–97.5th percentiles, i.e. the range containing the central 95\% of prompts in each split. Estimates use every question for splits with $n \leq 3,000$ and a uniform random sample of 3,000 for train. Two regimes are evident. The retrieval-rich splits — \texttt{train} (mean 2,836, 95\% band 1,841–3,484), \texttt{validation} (3,031; 2,052–3,451), \texttt{test} (2,894; 1,935–3,442) and \texttt{aljazeeraLate2025} (2,900; 1,905–3,482) — span roughly 1,800–3,500 tokens, whereas \texttt{aljazeera2026Q1} (632; 588–673) and \texttt{skysports2025} (610; 565–660) carry minimal retrieval context and are both 4.6\texttimes~shorter with far tighter spreads. The long-prompt splits dominate context budget; the two short-prompt OOD splits leave ample room and never approach generation-length limits.
  }
  \label{fig:data-audit_shared-qids}
\end{figure}

\begingroup
\setlength{\extrarowheight}{2pt}
\begin{longtable}{@{}p{0.81\linewidth}@{\hspace{1em}}>{\raggedright\arraybackslash}p{0.12\linewidth}@{}}
\caption{Representative exemplars from the evaluation datasets. For \texttt{OpenForesight} we
reproduce the full prompt verbatim; both shown questions are cases where retrieval returned no
articles, so the prompt is short. Most prompts additionally embed a block of retrieved news
passages (\texttt{Article 1}\,\ldots\,\texttt{5}) and are considerably longer. \texttt{OpenForesight}
rows are from the public \texttt{test} split (qids 645, 159); \texttt{MATH-500} from its \texttt{test}
split; AIME rows from \dataset{AIME24} and AMC rows from \dataset{AMC23} (the \dataset{AIME25} and
\dataset{AMC24} splits share an identical format).}
\label{tab:dataset-exemplars}\\
\toprule
\textbf{Prompt / Problem} & \textbf{Ground truth}\\
\midrule
\endfirsthead
\multicolumn{2}{@{}l}{\small\itshape Table~\ref{tab:dataset-exemplars} (continued)}\\
\toprule
\textbf{Prompt / Problem} & \textbf{Ground truth}\\
\midrule
\endhead
\midrule
\multicolumn{2}{r@{}}{\small\itshape continued on next page}\\
\endfoot
\bottomrule
\endlastfoot

\multicolumn{2}{@{}l}{\textbf{\dataset{OpenForesight} — forecasting} (\texttt{test} split)}\\[3pt]
{\scriptsize
You will be asked a forecasting question (which might be from the past). You have to come up with the best guess for the final answer. You will also be provided with a list of retrieved news articles summaries which you may refer to when coming up with your answer. Please provide your reasoning before stating your final answer and also express how likely you think your answer is to be correct (your confidence in your answer). \newline
Question Title: Where will the initial court hearing for the rapper charged with a terror offence be held? \newline
Question Background: A member of an Irish rap group has been charged with a terror offence and is awaiting his first court appearance. \newline
Resolution Criteria: \textless{}ul\textgreater{} \newline
\textless{}li\textgreater{}Official court documents or venue listings on the day of the hearing.\textless{}/li\textgreater{} \newline
\textless{}li\textgreater{}June 18, 2025, when the hearing takes place.\textless{}/li\textgreater{} \newline
\textless{}li\textgreater{}Official name of the court venue exactly as stated (e.g., 'City Court').\textless{}/li\textgreater{} \newline
\textless{}/ul\textgreater{} \newline
Expected Answer Type: string (location) \newline
Relevant passages from retrieved news articles: \newline
No relevant articles found. \newline
Think step by step about the information provided, reason about uncertainty and put your final answer (in the format asked) in \textless{}answer\textgreater{} \textless{}/answer\textgreater{} tags. You should also specify your confidence in your answer in \textless{}probability\textgreater{} \textless{}/probability\textgreater{} tags. The probability should be a number between 0 and 1. \newline
You will be rewarded based on the probability (p) you assign to your answer. Your answer will be evaluated using the BRIER SCORING RULE which is basically (- (1 - p)\textasciicircum{}2) if your answer is correct and (- 1 - p\textasciicircum{}2) if your answer is incorrect. For example, if p = 0.5, and your answer is incorrect, then your score will be (-1 - 0.5\textasciicircum{}2) = (-1 - 0.25) = -1.25 whereas if the answer was correct, then your score would be (- (1 - 0.5)\textasciicircum{}2) = (- (0.5)\textasciicircum{}2) = -0.25. Thus, the range of the score is [-2, 0] where your score lies between [-2, -1] if the answer is incorrect and [-1, 0] if the answer is correct. If your answer is correct, your will be REWARDED more if your probability is higher whereas if your answer is incorrect, your will be PENALIZED more if your probability is higher. YOU HAVE TO MAXIMIZE YOUR SCORE. \newline
Your final answer should be concise (NOT MORE THAN A FEW WORDS LONG) and your response SHOULD STRICTLY END with \textless{}answer\textgreater{} \textless{}/answer\textgreater{} tags and \textless{}probability\textgreater{} \textless{}/probability\textgreater{} tags.
} & {\scriptsize Westminster Magistrates' Court}\\
\addlinespace[4pt]
{\scriptsize
You will be asked a forecasting question (which might be from the past). You have to come up with the best guess for the final answer. You will also be provided with a list of retrieved news articles summaries which you may refer to when coming up with your answer. Please provide your reasoning before stating your final answer and also express how likely you think your answer is to be correct (your confidence in your answer). \newline
Question Title: Where will the second leg of the 2025 UEFA Champions League play-off tie between Rangers and Club Brugge be held? \newline
Question Background: In UEFA Champions League play-off ties, teams contest two legs, with each team hosting one match. \newline
Resolution Criteria: \textless{}ul\textgreater{} \newline
\textless{}li\textgreater{}\textless{}b\textgreater{}Source of Truth\textless{}/b\textgreater{}: UEFA official fixture list published on UEFA.com by August 19, 2025.\textless{}/li\textgreater{} \newline
\textless{}li\textgreater{}\textless{}b\textgreater{}Resolution Date\textless{}/b\textgreater{}: August 19, 2025.\textless{}/li\textgreater{} \newline
\textless{}li\textgreater{}\textless{}b\textgreater{}Accepted Answer Format\textless{}/b\textgreater{}: The full name of the stadium where the match will be played.\textless{}/li\textgreater{} \newline
\textless{}/ul\textgreater{} \newline
Expected Answer Type: string (stadium) \newline
Relevant passages from retrieved news articles: \newline
No relevant articles found. \newline
Think step by step about the information provided, reason about uncertainty and put your final answer (in the format asked) in \textless{}answer\textgreater{} \textless{}/answer\textgreater{} tags. You should also specify your confidence in your answer in \textless{}probability\textgreater{} \textless{}/probability\textgreater{} tags. The probability should be a number between 0 and 1. \newline
You will be rewarded based on the probability (p) you assign to your answer. Your answer will be evaluated using the BRIER SCORING RULE which is basically (- (1 - p)\textasciicircum{}2) if your answer is correct and (- 1 - p\textasciicircum{}2) if your answer is incorrect. For example, if p = 0.5, and your answer is incorrect, then your score will be (-1 - 0.5\textasciicircum{}2) = (-1 - 0.25) = -1.25 whereas if the answer was correct, then your score would be (- (1 - 0.5)\textasciicircum{}2) = (- (0.5)\textasciicircum{}2) = -0.25. Thus, the range of the score is [-2, 0] where your score lies between [-2, -1] if the answer is incorrect and [-1, 0] if the answer is correct. If your answer is correct, your will be REWARDED more if your probability is higher whereas if your answer is incorrect, your will be PENALIZED more if your probability is higher. YOU HAVE TO MAXIMIZE YOUR SCORE. \newline
Your final answer should be concise (NOT MORE THAN A FEW WORDS LONG) and your response SHOULD STRICTLY END with \textless{}answer\textgreater{} \textless{}/answer\textgreater{} tags and \textless{}probability\textgreater{} \textless{}/probability\textgreater{} tags.
} & {\scriptsize Jan Breydel Stadium}\\
\midrule

\multicolumn{2}{@{}l}{\textbf{\dataset{MATH-500} — math reasoning} (\texttt{test} split)}\\[3pt]
{\small Convert the point $(0,3)$ in rectangular coordinates to polar coordinates. Enter your answer in the form $(r,\theta),$ where $r > 0$ and $0 \le \theta < 2 \pi.$ \quad{\scriptsize(Precalculus, level 2)}}
 & {\small $\left( 3, \frac{\pi}{2} \right)$}\\
\addlinespace[4pt]
{\small If $f(x) = \frac{3x-2}{x-2}$, what is the value of $f(-2) +f(-1)+f(0)$? Express your answer as a common fraction. \quad{\scriptsize(Algebra, level 3)}}
 & {\small $\dfrac{14}{3}$}\\
\midrule

\multicolumn{2}{@{}l}{\textbf{\dataset{AIME24} — math reasoning}}\\[3pt]
{\small Let $x,y$ and $z$ be positive real numbers that satisfy the following system of equations: \[\log_2\!\left({x \over yz}\right) = {1 \over 2},\quad \log_2\!\left({y \over xz}\right) = {1 \over 3},\quad \log_2\!\left({z \over xy}\right) = {1 \over 4}.\] Then the value of $\left|\log_2(x^4y^3z^2)\right|$ is $\tfrac{m}{n}$ where $m$ and $n$ are relatively prime positive integers. Find $m+n$. \quad{\scriptsize(2024-II-4)}}
 & {\small $33$}\\
\addlinespace[4pt]
{\small Jen enters a lottery by picking $4$ distinct numbers from $S=\{1,2,3,\cdots,9,10\}.$ $4$ numbers are randomly chosen from $S.$ She wins a prize if at least two of her numbers were $2$ of the randomly chosen numbers, and wins the grand prize if all four of her numbers were the randomly chosen numbers. The probability of her winning the grand prize given that she won a prize is $\tfrac{m}{n}$ where $m$ and $n$ are relatively prime positive integers. Find $m+n$. \quad{\scriptsize(2024-I-4)}}
 & {\small $116$}\\
\midrule

\multicolumn{2}{@{}l}{\textbf{\dataset{AMC23} — math reasoning}}\\[3pt]
{\small Cities $A$ and $B$ are $45$ miles apart. Alicia lives in $A$ and Beth lives in $B$. Alicia bikes towards $B$ at 18 miles per hour. Leaving at the same time, Beth bikes toward $A$ at 12 miles per hour. How many miles from City $A$ will they be when they meet?}
 & {\small $27$}\\
\addlinespace[4pt]
{\small What is the degree measure of the acute angle formed by lines with slopes $2$ and $\frac{1}{3}$?}
 & {\small $45$}\\
\end{longtable}
\endgroup





\section{Temperature sweep}
\label{app:temp-sweep}

\begin{figure}[h]
  \centering
  \includegraphics[width=0.9\linewidth]{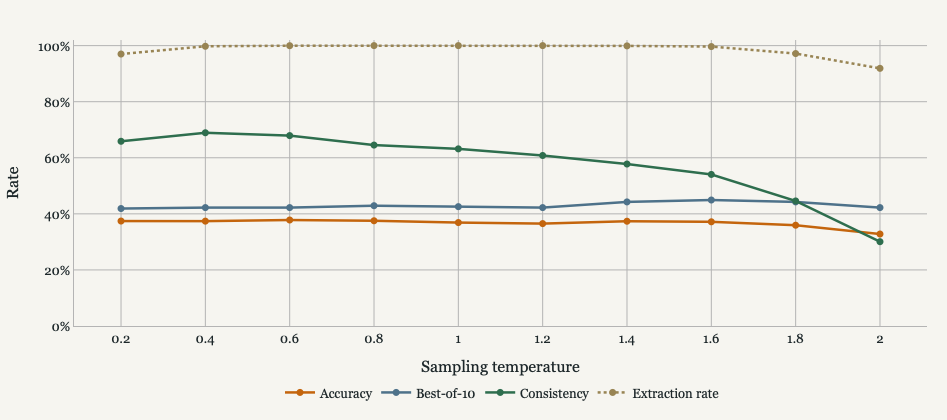}
  \caption{
  \textbf{Temperature sweep for \model{EF-8B} on the \dataset{OpenForesight-test}
  split.}
  Each curve summarizes $10$ rollouts per question across all $296$ test
  questions ($29{,}600$ generations) as a function of sampling temperature
  $T \in \{0.2, \dots, 2.0\}$. \emph{Self-consistency} (fraction of a question's
  rollouts in its modal answer cluster) falls monotonically from $65.9\%$
  ($T{=}0.2$) to $30.1\%$ ($T{=}2.0$), while \emph{per-rollout accuracy}
  (Claude-judged answer equivalence) holds flat at $37 \pm 1\%$ through
  $T{=}1.6$ before dropping to $32.8\%$ at $T{=}2.0$, and \emph{mean verbalized
  confidence} stays near $50\%$ throughout. The persistent gap between
  confidence (${\sim}50\%$) and accuracy (${\sim}37\%$) is the model's
  overconfidence, and it is largely temperature-invariant.
  }
  \label{fig:temp-sweep}
\end{figure}

Two regimes emerge (Fig.~\ref{fig:temp-sweep}). Across the moderate range
$T \lesssim 1.6$, correctness and confidence are essentially flat: per-rollout
accuracy holds at $37 \pm 1\%$ (a $1.3$\,pp spread, within the $\pm 0.9$\,pp
binomial standard error at $n = 2{,}960$), and mean verbalized confidence stays
near $50\%$ irrespective of $T$. Self-consistency, by contrast, declines
steadily across the whole sweep, from $66\%$ at $T = 0.2$ to $30\%$ at
$T = 2.0$, as the mean number of distinct answer clusters per question grows from
$1.86$ to $3.50$. Beyond $T \approx 1.6$ the model begins to emit malformed
outputs---answer-tag extraction falls from $100\%$ to $92\%$ at $T = 2.0$---and
accuracy degrades to $33\%$.

\begin{figure}[h]
  \centering
  \includegraphics[width=0.9\linewidth]{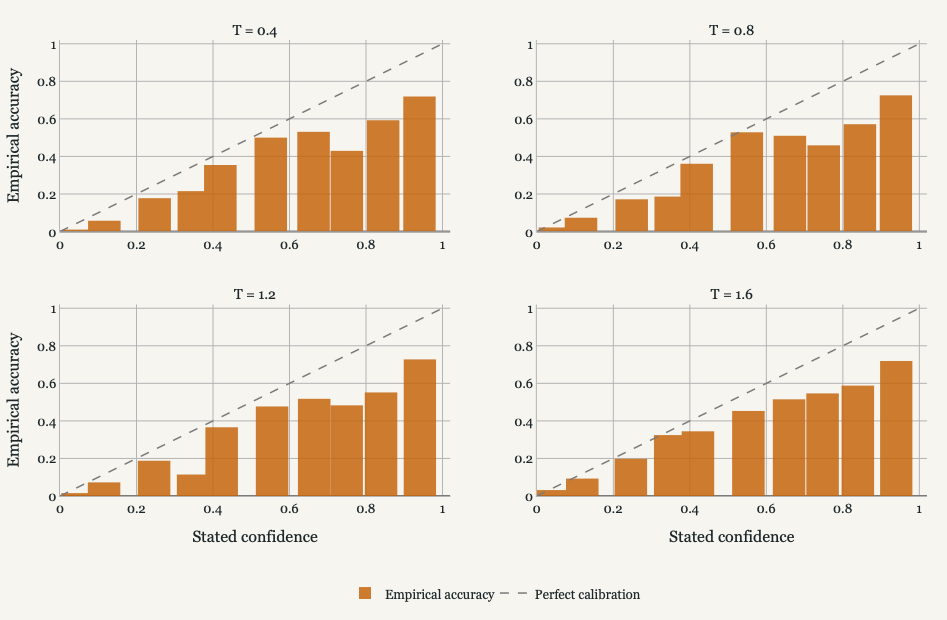}
  \caption{
  \textbf{Reliability of \model{EF-8B}'s verbalized confidence across temperature.}
  Reliability diagrams for $T = 0.4, 0.8, 1.2, 1.6$ on the
  \dataset{OpenForesight-test} split.
  In each panel the model's stated probabilities are grouped into ten
  equal-width confidence bins; bar height is the empirical accuracy within the
  bin and the dashed diagonal marks perfect calibration, so bars below the
  diagonal indicate overconfidence. The model is overconfident at every
  temperature, with probability mass concentrated in the high-confidence bins
  (${\sim}22\%$ of rollouts fall in the top bin $p > 0.9$, which resolves correct
  only ${\sim}72\%$ of the time).
  The shape is essentially temperature-invariant; calibration is mildly best at
  $T{=}1.6$. Per-panel expected calibration error: ECE $= 0.126, 0.118, 0.122,
  0.107$ for $T = 0.4, 0.8, 1.2, 1.6$ (Brier optimum $0.187$ at $T{=}1.2$; ECE
  optimum $0.107$ at $T{=}1.6$).
  }
  \label{fig:temp-reliability}
\end{figure}

The flatness of accuracy and confidence is reassuring for reproducibility, but
it exposes the calibration problem highlighted in the main text. Across the sweep
the model is consistently overconfident (ECE $0.11$--$0.15$, Brier
$0.19$--$0.21$; Fig.~\ref{fig:temp-reliability}), with probability mass
concentrated in the high-confidence bins and the reliability curve sitting below
the diagonal at every temperature.

\section{Probe training and sweep}
\label{app:probe-sweep}

We train the full grid of probes: every pooling architecture, at every layer, read at several points along the reasoning trace -- and report their metrics.
All probes are trained on the \dataset{OpenForesight-train} split (one \model{EF-8B} rollout per question, labeled for answer correctness by the LLM judge) and evaluated on \dataset{test} ($302$ questions $\times$ $10$ rollouts $= 3{,}020$), the same test set as the main-text result.

\paragraph{Sites.}
Each probe reads a pooled span ending at one of seven reasoning-anchored sites:
\texttt{initial} (through the opening \texttt{<think>}), the rollout fractions \texttt{r010}/\texttt{r030}/\texttt{r050}/\texttt{r070}/\texttt{r090}, and \texttt{final} (through the closing \texttt{</think>}). 
The \texttt{initial} span is shared by all rollouts of a question and therefore carries question-level signal only.

\paragraph{Training.}
Prompt $+$ truncated rollout, \texttt{AdamW} at learning rate \texttt{lr}~$= 10^{-3}$, batch size $8$, \texttt{MAX\_SEQ\_LEN} $8192$, seed $42$, temperature $T = 1.0$, covariance bottleneck $d_\mathrm{hidden}=64$, $10\%$ of the training questions held out for validation, and the checkpoint chosen at lowest validation loss (\emph{best-val} selection).
Each probe emits a single real-valued logit~$z$, mapped to a predicted probability of answer-correctness $p=\sigma(z)=1/(1+e^{-z})$, and is trained with the binary cross-entropy (BCE) loss -- the standard log-loss for a two-class probabilistic classifier. 
For a rollout with correctness label $y\in\{0,1\}$ ($y=1$ if the answer resolves correct),
\begin{equation}
  \mathcal{L}_\mathrm{BCE}(p,y) \;=\; -\big[\,y\log p + (1-y)\log(1-p)\,\big],
\end{equation}
averaged over the training set. 
The two branches penalize the two error modes: 
when $y=1$ the loss is $-\log p$ (large only if the probe assigned low probability to a correct answer) and when $y=0$ it is $-\log(1-p)$ (large only if it was confidently wrong). 
Minimizing it drives the probe's output toward the empirical probability of correctness, which is what makes the probe a \emph{calibrated} confidence estimate rather than a hard classifier. 


\paragraph{Metric.}
Each rollout contributes one prediction: the probe's raw probability $p=\sigma(z)$ (no recalibration) paired with its correctness label $y\in\{0,1\}$. 
All metrics are \emph{rollout-pooled} -- computed over the full set of rollouts pooled together, not averaged per question -- and follow the definitions of Section~\ref{sec:metrics}. 
We apply \emph{no temperature scaling}: best-val checkpoint selection yields calibrated heads directly, so the raw $p$ is the honest quantity, and AUROC is temperature-invariant in any case, so only ECE and Brier could be affected.

\begin{figure}[h]
  \centering
  \includegraphics[width=\linewidth]{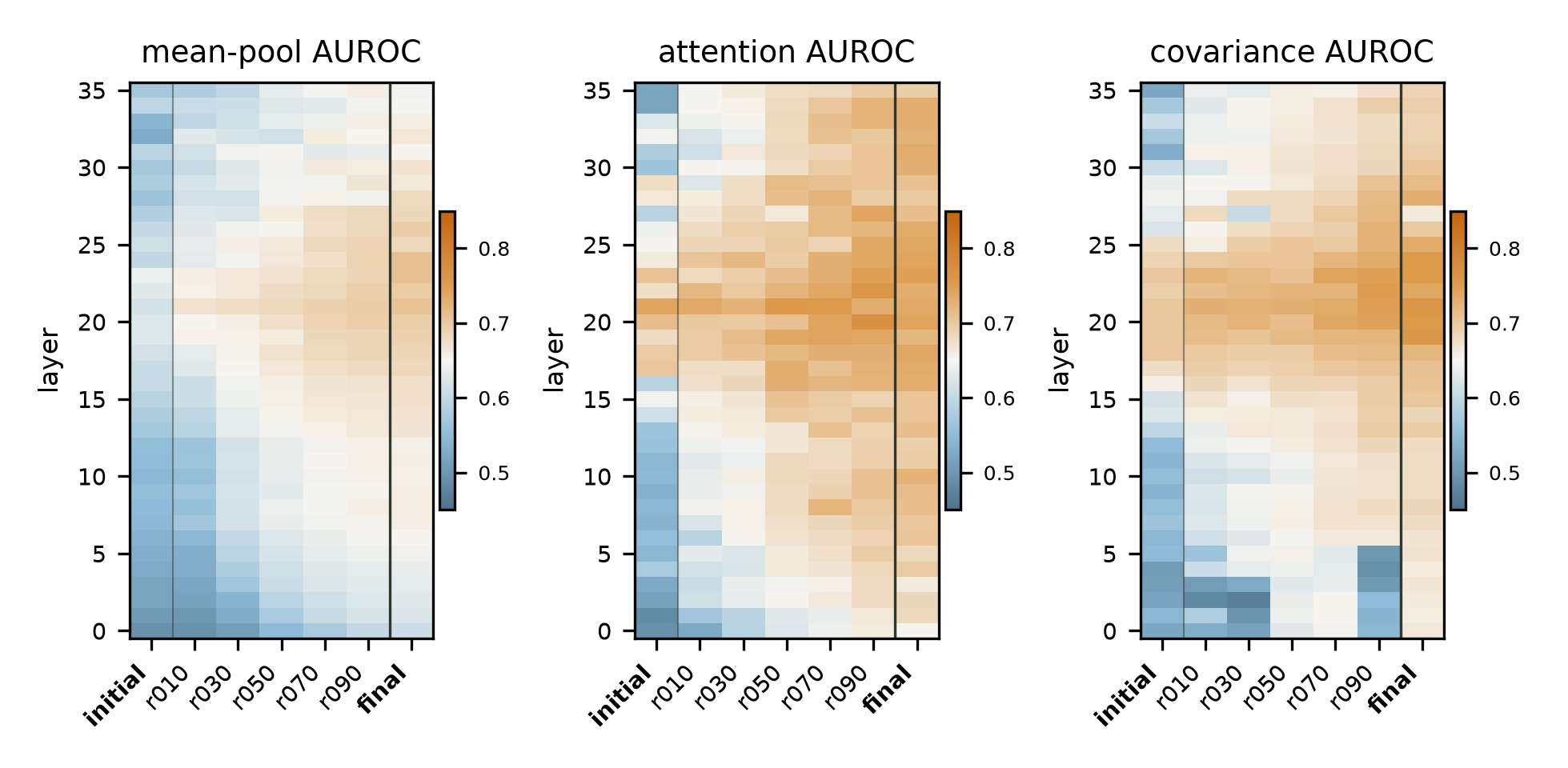}
  \caption{
\textbf{Probe AUROC across layers and reasoning sites}, 
for mean-pooling, attention, and covariance probes (left to right), evaluated on \dataset{test} ($N = 3{,}020$). 
Rows are the 36 layers of \model{EF-8B}; columns the seven sites, from \texttt{initial} through rollout fractions to \texttt{final}. Higher is better.
}
  \label{fig:probe-auroc}
\end{figure}
\begin{figure}[h]
  \centering
  \includegraphics[width=\linewidth]{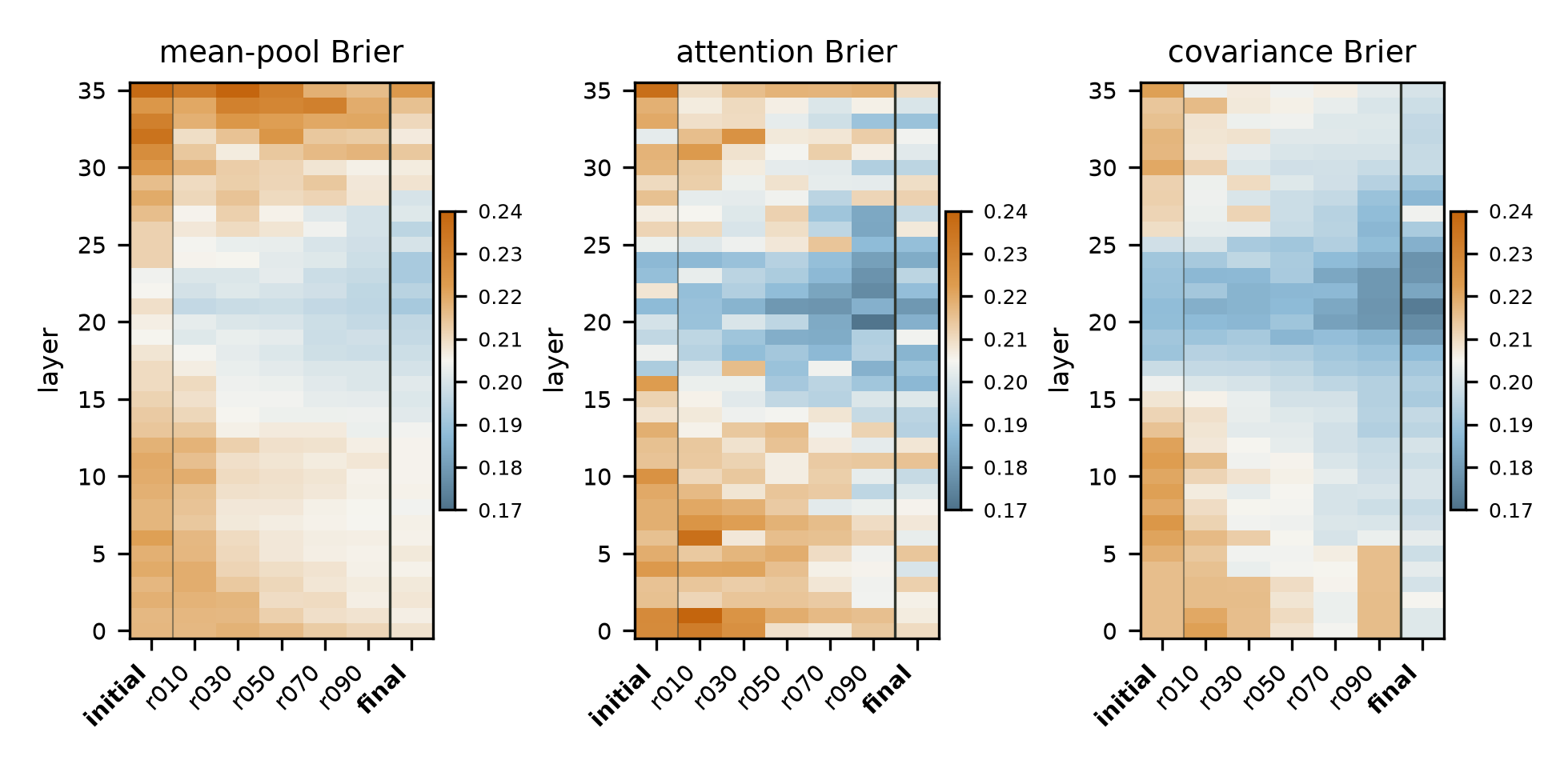}
  \caption{
\textbf{
Probe Brier score across layers and reasoning sites}, same probes, layout, and evaluation as Figure~\ref{fig:probe-auroc}. 
Cooler is lower (better); the low-Brier region mirrors the high-AUROC region.
\model{EF-8B}'s verbalized confidence achieves $\mathrm{Br} = 0.190$.
}
  \label{fig:probe-brier}
\end{figure}

\paragraph{Reading the grids.}
Figure~\ref{fig:probe-auroc} reports AUROC and Figure~\ref{fig:probe-brier} the Brier score (a more reliable single calibration summary than ECE, since a near-chance probe is trivially low-ECE). 
Across all three families, discrimination improves with reasoning depth, concentrates smoothly in the mid-to-late layers
($\sim$18--24), and is strongest for covariance pooling -- whose best cell is
layer~21/\texttt{final} -- which is why the deployed probe is a covariance probe.

\clearpage
\section{GLM Probe-Only Calibration Details}
\label{app:glm-probes}

The main GLM comparisons are reported in Table~\ref{tab:glm-probe-calibration} and Figures~\ref{fig:glm-probe-calibration-bars}, \ref{fig:glm-calibration}, and \ref{fig:glm-cross-model}.
This appendix records setup details, controls, and supplementary figures for the probe-only calibration experiments.

\paragraph{Models and data.}
We evaluate frozen \model{GLM-4.7-Flash} and \model{GLM-4.5-Air} models of two different sizes.
The dataset is another \dataset{OpenForesight} version whose question contexts were improved to be more relevant than the semantic-search contexts used in the OpenForecaster paper.
For each model, probes are trained on $11{,}835$ train rollouts ($3{,}945$ questions $\times$ 3 completions), selected on $1{,}930$ validation rollouts, and reported on $2{,}960$ held-out test rollouts.
The validation split is used for checkpoint, layer, and architecture selection; the test split is reserved for the headline numbers.

\paragraph{Probe input and training.}
The probe input is the prompt plus chain-of-thought, truncated before the final answer block.
Residual-stream activations are read from the frozen model, and only the probe weights are optimized.
Mean-pool linear probes use mask-aware pooled residual-stream vectors; attention and covariance probes use token-level activation sequences.
All probes use AdamW with learning rate $10^{-3}$ and weight decay $0.01$.
The GLM-4.7-Flash mean-pool probe is trained for up to 80 epochs with batch size 256 and early stopping patience 8; its sequence probes use batch size 32, up to 3 epochs, and patience 2.
The GLM-4.5-Air sweep follows the same recipe, with layer and architecture selection by validation cross-entropy.
ECE is computed on raw probabilities with no temperature scaling, Platt scaling, isotonic regression, or other recalibration.

\paragraph{Leakage controls and caveats.}
For \model{GLM-4.7-Flash}, the full test set often contains a drafted answer in the chain-of-thought before the final answer block.
The leakage-controlled comparison therefore uses the $423$ rollouts with both answer-clean reasoning and usable self-report, giving a probe-minus-verbalized-confidence AUROC delta of $+0.055$ with 95\% CI $[-0.006, 0.111]$.
For \model{GLM-4.5-Air}, the clean subset removes rollouts that leak a probability tag mid-reasoning, leaving $1{,}747$ rollouts and a clean AUROC delta of $+0.110$ with 95\% CI $[0.069, 0.149]$.
Thus, \model{GLM-4.7-Flash} should be read mainly as a raw-calibration result, while \model{GLM-4.5-Air} supports both ranking and calibration gains.

\paragraph{Controls and transfer notes.}
Shuffled-label probes sit near chance, and chain-of-thought-length baselines do not explain the GLM-4.5-Air signal.
For \model{GLM-4.7-Flash}, same-format validation-to-test and cross-format train-to-test probes are similar, which argues against the train/test context-format difference explaining the test result.
For \model{GLM-4.5-Air}, the best layer lies in a broad mid-stack plateau around layers 18--23; matched 24--40 comparisons reproduce the effect across \model{GLM-4.7-Flash} and \model{GLM-4.5-Air}.

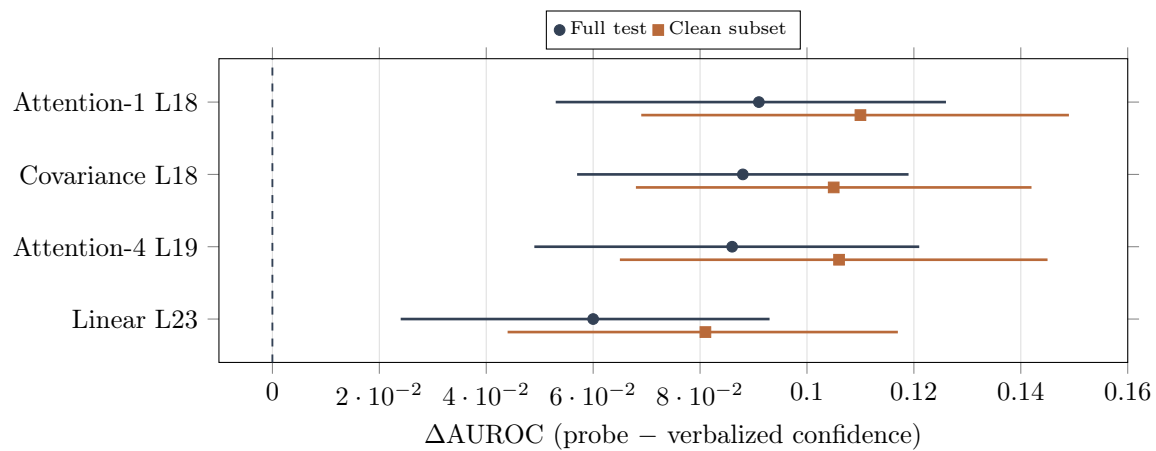
\begin{figure}[p]
\centering
\begin{tikzpicture}
\begin{axis}[
  width=0.82\linewidth,
  height=5.6cm,
  xmin=-0.01,
  xmax=0.16,
  ymin=0.4,
  ymax=4.6,
  xlabel={$\Delta$AUROC (probe $-$ verbalized confidence)},
  ytick={1,2,3,4},
  yticklabels={Linear L23, Attention-4 L19, Covariance L18, Attention-1 L18},
  xmajorgrids,
  major grid style={gray!25},
  tick align=outside,
  legend style={font=\scriptsize, at={(0.5,1.03)}, anchor=south, legend columns=2}
]
\draw[dashed, slate, line width=0.7pt] (axis cs:0,0.4) -- (axis cs:0,4.6);
\draw[slate, line width=1pt] (axis cs:0.024,1) -- (axis cs:0.093,1);
\draw[slate, line width=1pt] (axis cs:0.049,2) -- (axis cs:0.121,2);
\draw[slate, line width=1pt] (axis cs:0.057,3) -- (axis cs:0.119,3);
\draw[slate, line width=1pt] (axis cs:0.053,4) -- (axis cs:0.126,4);
\draw[ember, line width=1pt] (axis cs:0.044,0.82) -- (axis cs:0.117,0.82);
\draw[ember, line width=1pt] (axis cs:0.065,1.82) -- (axis cs:0.145,1.82);
\draw[ember, line width=1pt] (axis cs:0.068,2.82) -- (axis cs:0.142,2.82);
\draw[ember, line width=1pt] (axis cs:0.069,3.82) -- (axis cs:0.149,3.82);
\addplot[only marks, mark=*, mark size=2pt, slate] coordinates {(0.060,1) (0.086,2) (0.088,3) (0.091,4)};
\addplot[only marks, mark=square*, mark size=2pt, ember] coordinates {(0.081,0.82) (0.106,1.82) (0.105,2.82) (0.110,3.82)};
\legend{Full test, Clean subset}
\end{axis}
\end{tikzpicture}
\caption{\textbf{GLM-4.5-Air probe versus verbalized confidence.} Paired $\Delta$AUROC for each probe family at its best layer, comparing full test and leakage-controlled clean subsets. Bars are question-clustered 95\% bootstrap intervals.}
\label{fig:glm-probe-vs-selfreport}
\end{figure}

\begin{figure}[p]
\centering
\begin{tikzpicture}
\begin{axis}[
  width=0.88\linewidth,
  height=6.2cm,
  xmin=16,
  xmax=40,
  ymin=0.45,
  ymax=0.82,
  xtick={16,18,20,22,24,26,28,30,32,34,36,38,40},
  xlabel={Residual-stream layer},
  ylabel={Test AUROC},
  title={GLM-4.5-Air depth profile},
  grid=both,
  grid style={gray!20},
  tick align=outside,
  legend style={font=\scriptsize, at={(0.5,-0.24)}, anchor=north, legend columns=3},
  every axis plot/.append style={line width=0.9pt, mark size=1.2pt}
]
\addplot[draw=none, fill=goodfire2!13, forget plot] coordinates {(18,0.45) (23,0.45) (23,0.82) (18,0.82)} \closedcycle;
\addplot[slate, mark=*] coordinates {
(16,0.738)(17,0.742)(18,0.746)(19,0.749)(20,0.749)(21,0.750)(22,0.748)(23,0.751)(24,0.742)(25,0.732)(26,0.734)(27,0.729)(28,0.731)(29,0.732)(30,0.729)(31,0.727)(32,0.724)(33,0.724)(34,0.718)(35,0.717)(36,0.714)(37,0.717)(38,0.713)(39,0.713)(40,0.706)};
\addplot[ember, mark=square*] coordinates {
(16,0.764)(17,0.772)(18,0.783)(19,0.765)(20,0.770)(21,0.778)(22,0.779)(23,0.774)(24,0.762)(25,0.769)(26,0.743)(27,0.741)(28,0.740)(29,0.762)(30,0.757)(31,0.751)(32,0.749)(33,0.742)(34,0.753)(35,0.752)(36,0.721)(37,0.741)(38,0.717)(39,0.749)(40,0.749)};
\addplot[blue, mark=triangle*] coordinates {
(16,0.762)(17,0.761)(18,0.770)(19,0.777)(20,0.769)(21,0.776)(22,0.767)(23,0.770)(24,0.757)(25,0.761)(26,0.738)(27,0.741)(28,0.722)(29,0.732)(30,0.761)(31,0.744)(32,0.740)(33,0.753)(34,0.747)(35,0.743)(36,0.724)(37,0.717)(38,0.725)(39,0.744)(40,0.729)};
\addplot[forest, mark=diamond*] coordinates {
(16,0.758)(17,0.764)(18,0.779)(19,0.778)(20,0.778)(21,0.772)(22,0.771)(23,0.762)(24,0.754)(25,0.759)(26,0.756)(27,0.746)(28,0.749)(29,0.758)(30,0.746)(31,0.743)(32,0.743)(33,0.745)(34,0.743)(35,0.733)(36,0.733)(37,0.729)(38,0.729)(39,0.730)(40,0.728)};
\draw[dashed, goodfire1, line width=0.8pt] ({rel axis cs:0,0}|-{axis cs:16,0.691}) -- ({rel axis cs:1,0}|-{axis cs:16,0.691});
\draw[dotted, slate, line width=0.9pt] ({rel axis cs:0,0}|-{axis cs:16,0.490}) -- ({rel axis cs:1,0}|-{axis cs:16,0.490});
\legend{Linear, Attention-1, Attention-4, Covariance}
\end{axis}
\end{tikzpicture}
\caption{\textbf{GLM-4.5-Air depth profile.} Forecast-correctness decodability across residual-stream layers and probe families, showing a broad layer 18--23 plateau.}
\label{fig:glm-depth-profile}
\end{figure}
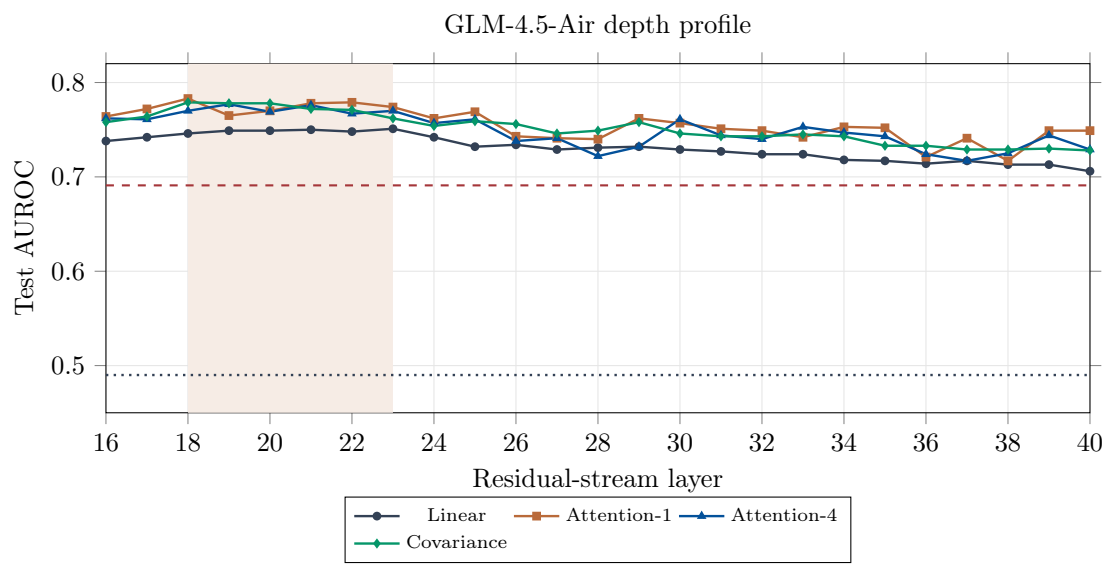

\clearpage
\section{OOD math probe stress test}
\label{app:math-probe}

The main OOD comparison is reported in Table~\ref{tab:ood-main}.
This appendix records the supporting setup.

\paragraph{Models and decode.}
We compare the untrained \model{Qwen3-8B} base model with our \model{Qwen3-8B} model trained according to the DCPO recipe of \citet{ma2026decouplingreasoningconfidenceresurrecting}.
The decode settings are matched across cells: temperature $0.7$, top-$p$ $0.8$, top-$k$ $20$, presence penalty $1.5$, maximum $3{,}000$ generated tokens, logprobs enabled, no thinking mode, and seed $42$.
The base model is evaluated with plain decoding; the DCPO-recipe model is evaluated with the verbal-confidence decode so that stated-confidence baselines are available.

\paragraph{Probe training and evaluation.}
Each probe is a one-stage binary correctness readout trained with BCE on frozen residual-stream activations.
The input is prompt plus reasoning text truncated before the first boxed answer, so the probe does not see the final answer or any trailing confidence line.
The selected sites are L18 last-token for the untrained base and L21 last-token for the DCPO-recipe model, chosen on held-out validation data and then frozen for evaluation.
We report AUROC and 15-bin ECE; deltas use qid-clustered bootstrap 95\% confidence intervals with 2{,}000 resamples.

\paragraph{Baselines and controls.}
The non-probe baselines are token-logprob confidence, isotonic-recalibrated token logprobs, raw verbalized confidence, isotonic-recalibrated verbalized confidence, and the decode-the-stated-number composite when the verbal decode is available.
Controls with random correctness labels and row-shuffled activations sit near chance, indicating that the real probe reads correctness structure from activations rather than a label or ordering artifact.
In-distribution diagnostics on clean \dataset{MATH-500} show that recalibrated verbalized confidence can outperform probes on ECE, so the main math result should be read as an OOD monitoring and discrimination result.
The DCPO probe-train pool overlaps the RLVR training distribution through \dataset{DeepScaleR}; therefore these results do not establish training-invariant correctness representations.

\begin{figure}[H]
\centering
\begin{tikzpicture}
\begin{axis}[
  width=0.7\linewidth, height=4.5cm, ybar, bar width=12pt,
  symbolic x coords={Base/plain, DCPO/verbal}, xtick=data,
  ymin=0.3, ymax=0.7, ylabel={AUROC}, title={Control probes (5 seeds, $\pm$ std)},
  enlarge x limits=0.5, legend style={font=\scriptsize, at={(0.5,-0.20)}, anchor=north, legend columns=2},
  ymajorgrids, major grid style={gray!30}]
\addplot[fill=slate, draw=slate, error bars/.cd, y dir=both, y explicit]
  coordinates {(Base/plain,0.473) +- (0,0.048) (DCPO/verbal,0.547) +- (0,0.027)};
\addplot[fill=forest, draw=forest, error bars/.cd, y dir=both, y explicit]
  coordinates {(Base/plain,0.501) +- (0,0.063) (DCPO/verbal,0.501) +- (0,0.062)};
\draw[dashed, gray] ({rel axis cs:0,0}|-{axis cs:Base/plain,0.5}) -- ({rel axis cs:1,0}|-{axis cs:Base/plain,0.5});
\legend{Random-label, Shuffled-activation}
\end{axis}
\end{tikzpicture}
\caption{
\textbf{Negative-control probes sit at chance.}
The dashed line marks AUROC $=0.5$.
Random-label and shuffled-activation probes are trained identically to the real probe, with 5 seeds each.
}
\label{fig:math-probe-controls}
\end{figure}
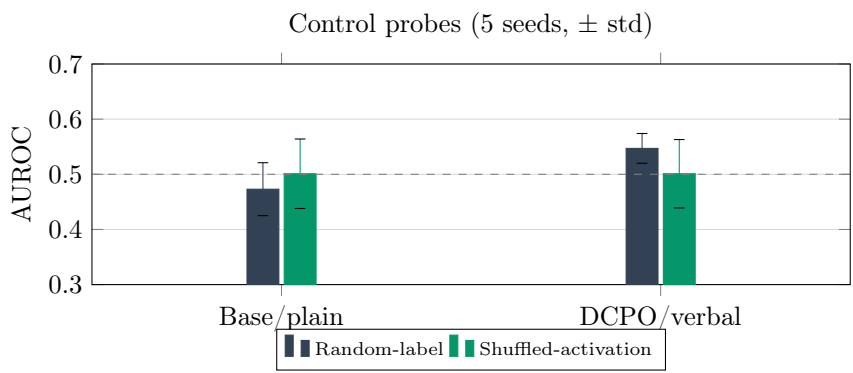

\end{document}